\pdfoutput=1

\documentclass[10pt,twocolumn,letterpaper]{article}

\usepackage[pagenumbers]{cvpr}

\usepackage{microtype}

\newcommand{\method}{MGOE}
\newcommand{\armstock}{VGGT}
\newcommand{\armft}{VGGT-ft}
\newcommand{\armaug}{Aug-only}
\newcommand{\armfilm}{\method(FiLM)}
\newcommand{\armbias}{\method(AttnBias)}
\newcommand{\armfilmbias}{\method(FiLM+AttnBias)}
\newcommand{\armrgb}{\armfilm{}+RGB}

\newcommand{\armgeo}{\armfilm{}+GeoPhoto}
\newcommand{\armbiasgeo}{\armbias{}+GeoPhoto}

\usepackage{bibunits}

\definecolor{cvprblue}{rgb}{0.21,0.49,0.74}
\usepackage[pagebackref,breaklinks,colorlinks,allcolors=cvprblue]{hyperref}
\usepackage{bibunits-backref-compat}

\def\confName{CVPR}
\def\confYear{2027}

\title{MGOE: Mask-Guided Object Exclusion for Feed-Forward 3D Reconstruction}
\author{Yuhang Zhang \qquad S\"{o}ren Schwertfeger\thanks{Corresponding author.}\\
ShanghaiTech University\\
{\tt\small \{zhangyh12024, soerensch\}@shanghaitech.edu.cn}
}

\begin{document}
\begin{bibunit}

\maketitle

\begin{abstract}
We introduce \method{}, a mask-guided framework that enables feed-forward 3D reconstruction to exclude user-specified objects and recover background geometry supported by other input views.
Dynamic objects can disrupt reconstruction, while even static objects may be unwanted in scene representations used by downstream applications.
Deleting unwanted points after reconstruction does not itself recover the occluded background or correct camera pose errors caused by object motion.
We train the model through geometry-guided synthetic occlusion, masking surfaces observed in other views while retaining their original geometric supervision.
We study feature-wise linear modulation (FiLM) and attention bias as alternative mask-conditioning mechanisms, yielding \armfilm{} and \armbias{}, respectively.
The FiLM variant yields stronger object exclusion on real dynamic scenes, reducing median background depth AbsRel from 0.2457 to 0.0257 and median foreground leakage from 84.01\% to 2.74\% on Bonn relative to a controlled fine-tuning baseline.
Evaluations on four clean-scene benchmarks further demonstrate that this configuration preserves general-purpose geometric reconstruction performance while improving selected metrics over a controlled fine-tuning baseline.
Code is available at \url{https://github.com/witsir/MGOE}.
\end{abstract}

\begin{figure*}[t]
\centering
\includegraphics[width=0.98\textwidth]{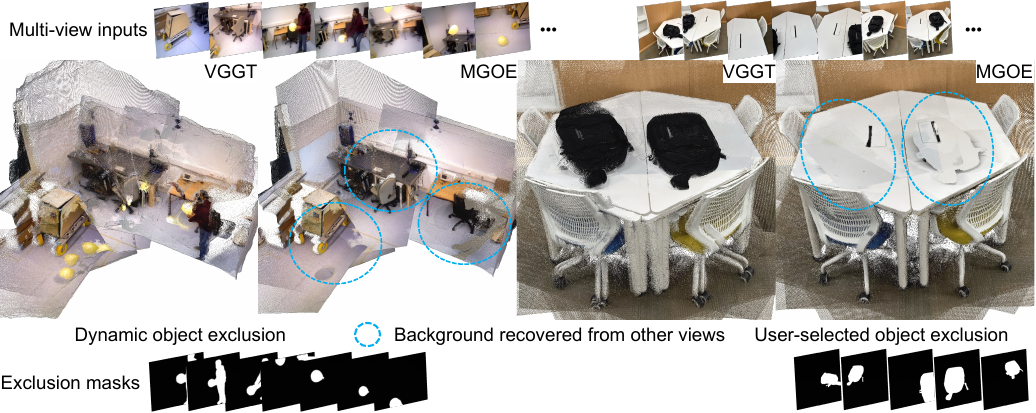}
\caption{\textbf{Mask-guided object exclusion for feed-forward 3D reconstruction.}
Compared with the original VGGT, \armfilm{} excludes dynamic objects (left) and a user-selected object (right) using exclusion masks, recovering background geometry from other input views.
The selected object changes position during capture, exposing the underlying surfaces in other views.}
\label{fig:teaser}
\end{figure*}

\section{Introduction}

Feed-forward 3D reconstruction predicts camera poses, depth maps, and point maps from multi-view images in a single forward pass, providing scene geometry for downstream applications.
The objects present during capture, however, are not necessarily the objects users want to retain in the reconstructed scene.
Dynamic objects can interfere with geometric estimation by corrupting cross-view correspondences, camera alignment, and background geometry~\cite{zhang2025monst3r,hu2026vggt4d,zhou2026page4d,he2026dynamicvggt,wimbauer2025anycam}.
Recent approaches mitigate this interference by separating dynamic from static content inside the reconstruction, through motion-aware segmentation, attention adaptation, or distractor-suppressing training~\cite{chen2025easi3r,goli2025romo,pan2026vgtw}.
Static objects that are only temporarily placed in the scene are likewise often unwanted in the reconstructed scene that downstream users need~\cite{he2019furniture,ye2024gaussiangrouping}.
This motivates a reconstruction process that allows users to specify which objects to exclude and recovers the background geometry behind them from other input views where it remains visible (\cref{fig:teaser}).

However, feed-forward reconstruction models such as VGGT~\cite{wang2025vggt} provide no explicit interface through which users can specify which image regions to exclude from reconstruction.
Reconstructed geometry can be post-processed by deleting the points of an unwanted instance~\cite{xia2025trainingfree} or filtering low-confidence predictions~\cite{hillemann2026uncertainty}, but these operations do not themselves recover the background geometry behind removed objects.
Prior work addresses missing scene content through generative synthesis~\cite{xia2025trainingfree} or scene-specific reconstruction that aggregates background observations across views~\cite{zhu2023occlusionfree}.
We incorporate user-specified exclusion masks early in the reconstruction pipeline, before unwanted observations propagate into geometric predictions.

We use scene geometry and cross-view visibility to guide synthetic occluder placement, retaining the original geometric targets to train the model to recover masked surfaces from other input views.
Explicit mask conditioning identifies the observations to exclude, allowing the model to apply this learned recovery to background surfaces behind user-specified objects at inference.
Users mark unwanted objects only where they appear and set the masks to zero elsewhere; when all masks are zero, we use standard feed-forward reconstruction.
Our training strategy does not rely on VGGT-specific architectural components or a particular mask-conditioning mechanism.
We instantiate and evaluate it on pretrained VGGT using FiLM and attention bias as alternative conditioning mechanisms.

On Bonn, \armfilm{} reduces background reconstruction error and foreground leakage relative to controlled fine-tuning and input-masking baselines.
Clean-scene evaluations further show that \armfilm{} preserves general-purpose geometric reconstruction performance when no exclusion is requested.

Our contributions are summarized as follows:
\begin{itemize}
    \item We introduce \method{}, a mask-guided feed-forward reconstruction framework that excludes user-specified objects and recovers background geometry observed in other input views.
    \item We develop a geometry-guided synthetic occlusion training strategy that masks cross-view observable surfaces while preserving geometric supervision, providing a surrogate task for learning object exclusion.
    \item We show that synthetic occlusion training already enables cross-view-supported recovery, while FiLM conditioning further improves background recovery and foreground exclusion.
\end{itemize}

\section{Related Work}

\subsection{Feed-Forward 3D Reconstruction}
VGGT, our backbone, jointly predicts camera parameters, point maps, depth maps, and point tracks from multi-view images \cite{wang2025vggt}.
CUT3R supports persistent-state online perception, while $\pi^3$ enables permutation-equivariant geometry prediction \cite{wang2025cut3r,wang2026pi3}.
To address dynamic scenes, MonST3R and D$^2$USt3R extend point-map reconstruction to dynamic geometry, while Easi3R exploits pretrained attention to disentangle camera and object motion \cite{zhang2025monst3r,han2025d2ust3r,chen2025easi3r}.
More directly, VGGT4D and PAGE-4D exploit motion-aware features to separate dynamic and static regions, whereas DynamicVGGT explicitly models temporally coherent point motion \cite{hu2026vggt4d,zhou2026page4d,he2026dynamicvggt}.
VGGT-$\Omega$ further scales feed-forward reconstruction to improve performance on both static and dynamic scenes~\cite{wang2026vggtomega}.
RobustVGGT addresses a complementary issue by rejecting geometrically inconsistent input views rather than suppressing within-view dynamic objects~\cite{han2026robustvggt}.
Despite these advances, explicitly removing distractors while recovering the underlying static geometry remains a distinct challenge.

\subsection{Object Removal and Scene De-occlusion}

Pose-calibrated, scene-specific NeRF and 3DGS methods address object removal through multi-view segmentation and inpainting; some exploit cross-view depth cues to improve geometric consistency \cite{mirzaei2023spinnerf,liu2024deocclusion,ye2024gaussiangrouping,wang2024gscream,huang2025dgic}.
Other scene-specific approaches recover occlusion-free scenes by aggregating background observations across viewpoints~\cite{zhu2023occlusionfree}.
Generative methods, including FoundationDeOcclusion, use diffusion priors to synthesize occluded appearance or geometry \cite{hu2024inout,shi2025occludenerf,lai2025diffusiongs,chen2026fast}.
However, surfaces unseen in every input view remain prior-driven estimates, not verifiable recovery of the true scene geometry.
Feed-forward methods such as DGGS, GenWildSplat, and Wild3R suppress transients in generalizable 3DGS pipelines; their primary 3D output is a Gaussian scene representation rather than explicit per-pixel point maps \cite{liao2026dggs,gupta2026genwildsplat,furutani2026wild3r}.
VGTW predicts depth and point maps while suppressing transient distractors, rather than targeting user-selected static objects~\cite{pan2026vgtw}.
FreeInpaint addresses reference-based 3D inpainting with generative support~\cite{pan2026freeinpaint}.
Our focus is user-specified object exclusion with background geometry recovered from actual observations in other input views.

\begin{figure*}[t]
\centering
\includegraphics[width=0.98\textwidth]{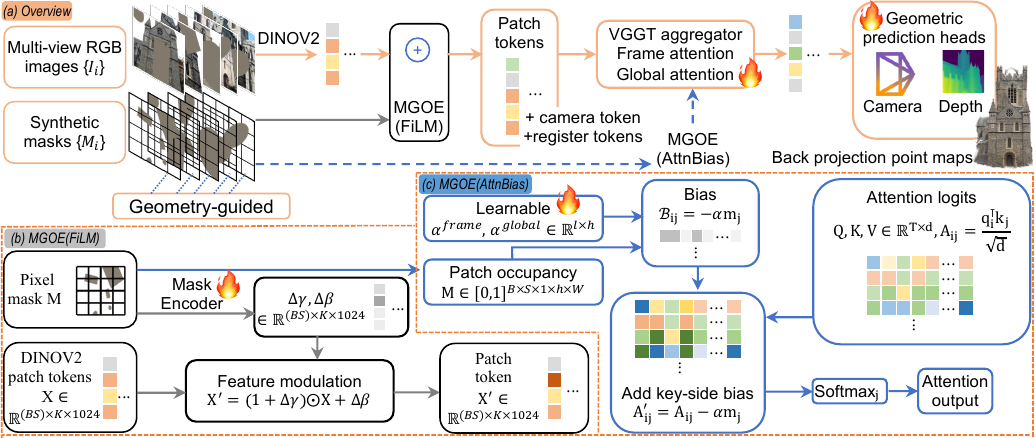}
\caption{\textbf{\method{} architecture.}
Given multi-view images and object masks, \method{} extends the VGGT pipeline with two alternative mask-conditioning mechanisms.
At inference, synthetic training masks are replaced by user-provided object masks.
(a) Overview of the shared feed-forward reconstruction pipeline.
(b) \armfilm{} modulates DINOv2 patch tokens before the VGGT Aggregator.
Zero initialization makes the modulation initially identity for any mask; bias-free projections keep it identity for empty masks throughout training.
(c) \armbias{} introduces learnable, layer- and head-specific key-side biases into Frame and Global Attention, suppressing masked keys while retaining masked queries for cross-view information aggregation.}
\label{fig:architecture}
\end{figure*}

\begin{figure}[t]
\centering
\includegraphics[width=\linewidth]{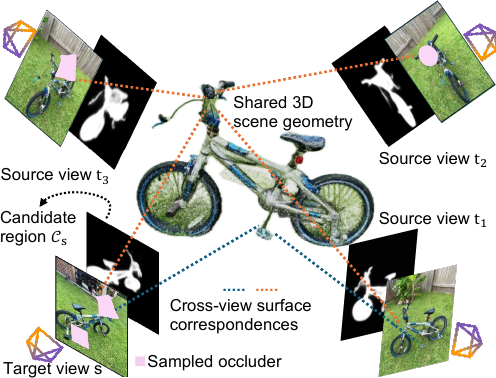}
\caption{\textbf{Geometry-guided synthetic occlusion.}
Cross-view surface correspondences define candidate regions $\mathcal{C}_s$ for occluder placement in target view $s$.
Pink shapes denote sampled occluders; colored dotted lines connect corresponding surface points.
We mask the sampled regions while retaining their original geometric targets, training the model to recover geometry using observations from other views.}
\label{fig:synthetic-occlusion}
\end{figure}

\subsection{Synthetic Occlusion}

CroCo masks one view for cross-view geometric pretraining, while UASNet applies asymmetric occlusion augmentation to stereo inputs for disparity estimation \cite{weinzaepfel2022croco,mao2021uasnet}.
ZeroShape-W and Amodal3R train on synthetically occluded object images, while GENA3D generates 3D-consistent occlusions across rendered views \cite{cho2025robust,wu2025amodal3r,zhou2026gena3d}.
DGGS and GenWildSplat introduce synthetic distractors into multi-view scene training \cite{liao2026dggs,gupta2026genwildsplat}.
\method{} uses ground-truth depth and camera poses to guide occlusion placement toward cross-view-observable surfaces, while retaining their original depth supervision.

\section{Method}

The geometry-guided synthetic occlusion strategy of \method{} does not depend on a specific mask-conditioning mechanism.
We instantiate mask conditioning in pretrained VGGT using FiLM as the main configuration, \armfilm{}, and evaluate key-side attention bias as an alternative, \armbias{}.
\method{} takes multi-view images and exclusion masks as input and predicts camera parameters and depth maps, from which point maps are obtained by back-projection (\cref{fig:architecture}).
In our main FiLM configuration, we mean-fill the excluded regions, encode the images with DINOv2~\cite{oquab2024dinov2}, and modulate the patch tokens with the masks before VGGT's alternating-attention backbone aggregates information across views.
We retain VGGT's camera and depth heads and adopt their losses from its public training implementation (Supplementary \cref{sec:supp-heads-losses}).

\subsection{Geometry-Guided Synthetic Occlusion}

We train \method{} to recover geometry from other views by synthetically hiding image regions while retaining their original geometric targets.
Training datasets provide geometric validity masks and may require existing dynamic objects to be masked before augmentation.
Unrestricted random masking ignores these constraints and can select regions lacking valid targets or usable cross-view observations, whose availability varies with input-view count and overlap.
For each input image $I_s$, $s\in\{1,\ldots,S\}$, let $V_s$ mark valid ground-truth geometry and $B_s$ mark dataset-required exclusions.
Using ground-truth depth and camera poses, we identify candidate regions $\mathcal{C}_s$ with valid geometry and non-excluded observations in both the target and another input view (\cref{fig:synthetic-occlusion}).
We sample occluder centers uniformly from these regions, falling back to uniform full-image sampling when a candidate region is empty.
The resulting synthetic mask $P_s$ is combined with $B_s$ to form the final exclusion mask $M_s$ and masked input $\widetilde I_s$:
\begin{equation}
\begin{aligned}
M_s(p) &= \max\{B_s(p),P_s(p)\},\\
\widetilde I_s(p) &=
\begin{cases}
I_s(p), & M_s(p)=0,\\
\boldsymbol{\mu}_{\mathrm{ImageNet}}, & M_s(p)=1.
\end{cases}
\end{aligned}
\label{eq:synthetic-masked-input}
\end{equation}
Here, $p\in\Omega_s$ indexes the $H\times W$ image domain; all masks are binary, with one indicating membership.
The ImageNet mean $\boldsymbol{\mu}_{\mathrm{ImageNet}}$ yields zero-valued RGB inputs after normalization.

Masking changes the observations but preserves the geometric supervision domain:
\begin{equation}
\mathcal{G}_s=\{p\in\Omega_s\mid V_s(p)=1\}.
\label{eq:geometric-supervision-domain}
\end{equation}
Thus, valid masked pixels remain supervised whether or not they retain support in another input view, and camera targets remain unchanged.
Cross-view visibility guides occluder placement rather than selecting geometric supervision pixels.
This task trains recovery of observed surfaces hidden by synthetic masks as a surrogate for recovering background behind user-masked objects at inference.
Supplementary \cref{sec:supp-synthetic-occlusion} specifies the visibility test, occluder sampling, coverage control, and mask-regeneration procedure.

\subsection{Mask Conditioning}
\label{sec:mask_conditioning}

We condition the VGGT alternating-attention backbone on the exclusion masks $\{M_s\}_{s=1}^{S}$ to limit the propagation of unreliable observations.
During training, conditioning uses the combined exclusion masks $M_s$; at inference, users provide masks for the objects to exclude.
We use FiLM~\cite{perez2018film} in \armfilm{} and key-side attention bias in \armbias{}.

\paragraph{FiLM.}
FiLM uses the exclusion mask to adjust image features before information is exchanged across views.
For view $s$, let $\mathbf{F}_s\in\mathbb{R}^{N\times C}$ contain the $N$ DINOv2 patch tokens, each with $C$ feature channels.
A lightweight MaskEncoder converts $M_s$ into a scale adjustment $\Delta\boldsymbol{\gamma}_s$ and a shift $\Delta\boldsymbol{\beta}_s$, both of shape $N\times C$:
\begin{equation}
    \widetilde{\mathbf{F}}_s
    = (\mathbf{1}+\Delta\boldsymbol{\gamma}_s)
      \odot\mathbf{F}_s + \Delta\boldsymbol{\beta}_s,
    \label{eq:mask_film}
\end{equation}
where $\mathbf{1}$ is the all-ones tensor and $\odot$ denotes element-wise multiplication.
Zero-initialized mask projections leave all patch features unchanged at initialization.
Because these projections have no bias, an empty mask produces zero modulation regardless of the learned weights.

\paragraph{Attention bias.}
The attention-bias mechanism uses the mask to reduce the attention scores assigned to excluded image regions.
For each key token $j$, let $m_j\in[0,1]$ be the fraction of masked pixels in its patch; camera and register tokens have $m_j=0$.
We subtract a penalty proportional to this fraction from the query--key similarity before softmax:
\begin{equation}
    L_{ij}^{a,\ell,h}
    = \frac{\mathbf{q}_i^{\top}\mathbf{k}_j}{\sqrt{d}}
      - \alpha_{a,\ell,h}m_j,
    \label{eq:attention_bias}
\end{equation}
Here, $\mathbf{q}_i,\mathbf{k}_j\in\mathbb{R}^{d}$ are the query and key vectors for tokens $i$ and $j$ in the current attention block.
The learned nonnegative coefficient $\alpha_{a,\ell,h}$ controls the penalty separately for attention type $a\in\{\mathrm{frame},\mathrm{global}\}$, layer $\ell$, and head $h$; these block indices are omitted from $\mathbf{q}_i$ and $\mathbf{k}_j$ for readability.
More heavily masked patches receive larger penalties, while masked tokens remain available as queries that gather information from other observations.
Initializing $\alpha_{a,\ell,h}$ to zero preserves the original attention operation at initialization, and an empty mask makes the penalty zero throughout training.
Both mechanisms retain all tokens and the camera/depth heads.
Supplementary \cref{sec:film_conditioning,sec:attention_bias_conditioning} provide the mask-projection, patch-pooling, and coefficient-parameterization details.

\begin{table*}[t]
\centering
\caption{\textbf{Object exclusion on Bonn.}
All entries use the same 20 videos, ten target frames, recovery region $R_{10}$, and foreground evaluation set.
Only the number of input views changes across columns.
Leakage is a fraction; ATE uses Sim(3) alignment on the ten target frames.
Depth scale is fitted independently per model, video, input condition, and view count using unmasked control pixels in the ten targets; object and box masks yield different control regions.}
\label{tab:bonn-exclusion}
\footnotesize
\setlength{\tabcolsep}{3pt}
\begin{tabular}{@{}llrrrrrrrrr@{}}
\toprule
& & \multicolumn{3}{c}{Background AbsRel $\downarrow$} & \multicolumn{3}{c}{Foreground leakage $\downarrow$} & \multicolumn{3}{c}{Pose ATE (m) $\downarrow$} \\
Model & Input fill & 10 views & 16 views & 24 views & 10 views & 16 views & 24 views & 10 views & 16 views & 24 views \\
\midrule
\armstock{} & Raw & 0.3898 & 0.3891 & 0.3891 & 0.9833 & 0.9871 & 0.9883 & 0.0492 & 0.0494 & 0.0459 \\
\armstock{} & Object mask, mean & 0.4043 & 0.4005 & 0.3919 & 0.9898 & 0.9853 & 0.9892 & 0.0420 & 0.0383 & 0.0376 \\
\armstock{} & Object mask, black & 0.3312 & 0.4123 & 0.4069 & 0.8912 & 0.9335 & 0.9288 & 0.0440 & 0.0415 & 0.0387 \\
\armstock{} & Object mask, white & 0.4310 & 0.4396 & 0.4411 & 0.9897 & 0.9926 & 0.9881 & 0.0424 & 0.0382 & 0.0363 \\
\armstock{} & Box mask, mean & 0.0869 & 0.0801 & 0.0800 & 0.4766 & 0.4312 & 0.4808 & 0.0467 & 0.0433 & 0.0426 \\
\armstock{} & Box mask, black & 0.0530 & 0.0505 & 0.0515 & 0.1559 & 0.1574 & 0.1764 & 0.0464 & 0.0448 & 0.0431 \\
\armstock{} & Box mask, white & 0.0970 & 0.0883 & 0.0860 & 0.5680 & 0.5111 & 0.5400 & 0.0456 & 0.0416 & 0.0411 \\
\midrule
\armft{} & Object mask, mean & 0.2457 & 0.2578 & 0.2643 & 0.8401 & 0.8946 & 0.8920 & 0.0376 & 0.0347 & 0.0351 \\
\armaug{} & Object mask, mean & 0.0632 & 0.0661 & 0.0699 & 0.3157 & 0.3073 & 0.4182 & 0.0327 & 0.0326 & 0.0325 \\
\armbias{} & Object mask, mean & 0.0612 & 0.0547 & 0.0630 & 0.3684 & 0.2412 & 0.3898 & 0.0333 & 0.0299 & 0.0303 \\
\armfilm{} & Object mask, mean & \textbf{0.0257} & \textbf{0.0248} & \textbf{0.0252} & \textbf{0.0274} & \textbf{0.0286} & \textbf{0.0302} & \textbf{0.0295} & \textbf{0.0298} & \textbf{0.0292} \\
\bottomrule
\end{tabular}
\end{table*}

\subsection{Cross-View Appearance Recovery}
\label{sec:cross-view-fusion}

To improve the appearance of the reconstructed point cloud, we perform post-hoc cross-view color transfer using the predicted geometry.
For view $i$, let $\widehat D_i$ denote predicted depth, $K_i\in\mathbb{R}^{3\times3}$ predicted intrinsics, and $T_i=[\mathbf R_i\mid\mathbf t_i]\in\mathbb{R}^{3\times4}$ the predicted camera-from-world transform, consistent with the convention in Supplementary \cref{sec:supp-synthetic-occlusion}.
For a masked pixel $p=(u,v)$ with valid predicted depth, let $\tilde p=(u,v,1)^{\top}$ be its homogeneous image coordinates.
We unproject it to a world point $\mathbf X_w\in\mathbb{R}^3$ and project it into each other input view $j$:
\begin{equation}
\begin{aligned}
    \mathbf X_w
    &= \mathbf R_i^{\top}
       \bigl(\widehat D_i(p)K_i^{-1}\tilde p-\mathbf t_i\bigr),\\
    q_{ij}
    &= \pi\!\left(K_jT_j
       \begin{bmatrix}\mathbf X_w\\1\end{bmatrix}\right),
\end{aligned}
\end{equation}
where $\pi(x,y,z)=(x/z,y/z)$ performs perspective division and $q_{ij}$ is the source image location.
A source observation is accepted only when its projection is in bounds, its four bilinear depth neighbors are valid and unmasked, and the projected depth agrees with the sampled source depth within a tolerance.
Among valid sources, we select the smallest depth residual, using confidence only to break ties.
The corresponding RGB is transferred to the target pixel; otherwise, its input color is retained.
This post-processing changes only point-cloud colors, leaving XYZ positions, camera poses, and point counts unchanged.
See Supplementary \cref{sec:supp-bonn-qualitative} for additional qualitative visualizations.

\section{Experiments}
\label{sec:experiments}

\subsection{Experimental Setup}
\label{sec:experimental-setup}

\paragraph{Training and comparisons.}
All fine-tuned variants share VGGT initialization and a two-stage, 19,200-step schedule on CA-1M~\cite{lazarow2025cubify}, ScanNet++~\cite{yeshwanth2023scannetpp}, Hypersim~\cite{roberts2021hypersim}, BlendedMVS~\cite{yao2020blendedmvs}, KITTI-360~\cite{liao2023kitti360}, and Co3Dv2~\cite{reizenstein2021co3d}.
We compare pretrained VGGT (\armstock{}), controlled fine-tuning (\armft{}), synthetic occlusion training without mask conditioning (\armaug{}), and the two mask-conditioned models, \armfilm{} and \armbias{}.
The \armft{} baseline disables additional synthetic occlusion and mask conditioning while retaining dataset-required exclusions, separating adaptation to our training protocol from the effects of our method.
Training details and computational cost are provided in Supplementary \cref{sec:supp-training-protocol,sec:supp-compute}.

\paragraph{Inputs and evaluation.}
For object exclusion, excluded RGB pixels receive the mean color in \cref{eq:synthetic-masked-input} unless otherwise specified; only mask-conditioned models additionally receive exclusion masks.
Clean-input evaluations use original RGB and all-zero conditioning masks.
We evaluate object exclusion on Bonn~\cite{palazzolo2019refusion} using controlled comparisons among VGGT-based variants.
Separately, we compare against external methods on Bonn and TUM RGB-D~\cite{sturm2012benchmark}, restricting these comparisons to camera pose estimation (\cref{sec:external-pose}).
Bonn/TUM object masks come from SAM3~\cite{carion2026sam3} tracking and manual object selection, with offline ground-truth geometry assisting Bonn selection.
Within each dataset, controlled VGGT-based comparisons share mask preprocessing; dataset-specific processing and external-method inputs are detailed in Supplementary \cref{sec:supp-external-pose}.
We evaluate clean geometry on RealEstate10K (RE10K)~\cite{zhou2018stereo}, ScanNet~\cite{dai2017scannet}, DTU~\cite{jensen2014large}, and ETH3D~\cite{schops2017multi}.
A controlled synthetic diagnostic examines supported recovery and foreground residuals.

\subsection{Object Exclusion in Real Dynamic Scenes}
\label{sec:object-exclusion}

\begin{figure*}[t]
\centering
\includegraphics[width=0.98\textwidth]{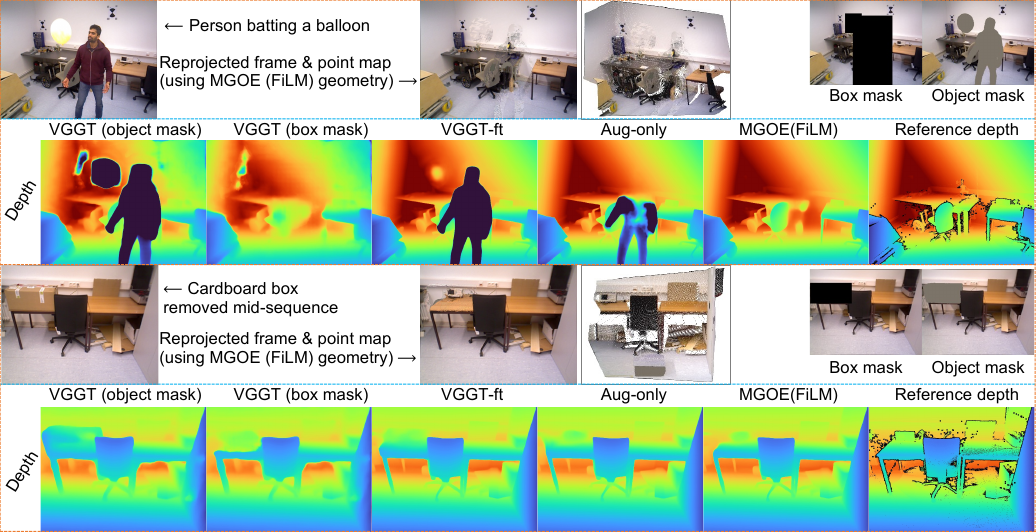}
\caption{\textbf{Qualitative object exclusion on Bonn.}
\armfilm{} suppresses dynamic and temporarily placed foreground objects while recovering background geometry observed in other frames.
Object masks use mean filling; the VGGT box-mask baseline uses black filling.}
\label{fig:bonn-qualitative}
\end{figure*}

\paragraph{Protocol and baselines.}
We evaluate 20 Bonn dynamic videos against static-background reference geometry, using 10, 16, and 24 input views while scoring the same ten target frames and fixed evaluation regions.
Background absolute relative depth error (AbsRel) measures recovery on $R_{10}$, the excluded target pixels with valid background geometry supported by another view in the ten-view input set.
Foreground leakage is the fraction of evaluated foreground pixels whose predicted depth lies in front of the static background beyond a tolerance.
Pose absolute trajectory error (ATE) uses Sim(3) alignment on the ten targets.
\Cref{tab:bonn-exclusion} reports medians across videos.

To test whether RGB filling alone enables exclusion, we compare pretrained VGGT with object and box masks, each filled with mean, black, or white color.
Box masks also remove the object-shaped boundary and surrounding context.
All fine-tuned variants use identical object masks and mean-filled RGB.
Black-filled boxes give the lowest background error and leakage among stock filling baselines, but their ten-view AbsRel (0.0530) and leakage (0.1559) exceed those of \armfilm{} (0.0257 and 0.0274).
Evaluation details and mask-filling analysis are provided in Supplementary \cref{sec:supp-bonn-protocol,sec:supp-mask-filling}.

\paragraph{Recovery and foreground exclusion.}
At ten views, synthetic occlusion training reduces background AbsRel from 0.2457 (\armft{}) to 0.0632 (\armaug{}), and adding FiLM conditioning in \armfilm{} further reduces it to 0.0257.
Foreground leakage decreases from 0.8401 to 0.3157 and then to 0.0274 along the same comparison.
\armbias{} also improves substantially over \armft{}, but has higher recovery error and leakage than \armfilm{} at each input budget.
\armfilm{} also achieves the lowest median ATE among the displayed conditions at each input budget, reaching 0.0295\,m at ten views compared with 0.0376\,m for \armft{}.
Increasing the input-view count does not yield uniformly monotonic improvements on these fixed targets.
\Cref{fig:bonn-qualitative} illustrates dynamic and stationary object exclusion; further error maps and geometry visualizations are provided in Supplementary \cref{sec:supp-bonn-qualitative}.
For qualitative object exclusion in more real-world scenes, see Supplementary \cref{sec:supp-qualitative-exclusion}.

\subsection{Camera Pose Estimation in Dynamic Scenes}
\label{sec:external-pose}

\begin{table}[ht]
\centering
\caption{\textbf{Camera pose comparison on 24-view inputs.}
ATE in meters after Sim(3) alignment over all 24 input frames.
Raw denotes unmodified RGB; Masked denotes mean-filled RGB with an object mask supplied to \armfilm{}.
Bonn reports the median over 20 dynamic videos; TUM reports the median of seven sequence medians, each computed over its motion windows (26 windows total).}
\label{tab:external-pose}
\small
\setlength{\tabcolsep}{4pt}
\begin{tabular}{llrr}
\toprule
Method & Input & Bonn $\downarrow$ & TUM $\downarrow$ \\
\midrule
MoRe-FA~\cite{fang2026more} & Raw & 0.0552 & 0.0155 \\
Easi3R-DUSt3R~\cite{chen2025easi3r} & Raw & 0.0701 & 0.0354 \\
MonST3R-GO~\cite{zhang2025monst3r} & Raw & 0.1154 & 0.0958 \\
\armstock{} & Raw & 0.0394 & 0.0115 \\
\armfilm{} & Raw & 0.0298 & 0.0126 \\
\midrule
\armfilm{} & Masked & 0.0264 & 0.0142 \\
\bottomrule
\end{tabular}
\end{table}

Camera pose provides a common evaluation target for user-specified object exclusion and dynamic-scene reconstruction.
MoRe-FA uses feed-forward inference, whereas Easi3R-DUSt3R and MonST3R-GO use flow-based global alignment with temporal pair graphs~\cite{fang2026more,chen2025easi3r,zhang2025monst3r}.
\method{} accepts nonconsecutive views; we evaluate all methods on identical 24-view Bonn and TUM samples in their sampled order.
This shared sampling protocol includes nonchronological Bonn inputs, so the results characterize pose estimation under these input conditions.
With raw RGB and an all-zero conditioning mask, \armfilm{} has lower median ATE than the three external methods on both datasets, while pretrained VGGT has the lowest TUM ATE (\cref{tab:external-pose}).
Providing object masks lowers the displayed Bonn median and raises the TUM median; paired comparisons do not establish a consistent pose improvement.
Supplementary \cref{sec:supp-external-pose} details the input conditions, method configurations, and scoring protocol.

\subsection{Controlled Synthetic Occlusion Diagnostic}
\label{sec:cross-view-recovery}

\begin{figure}[ht]
\centering
\includegraphics[width=0.98\columnwidth]{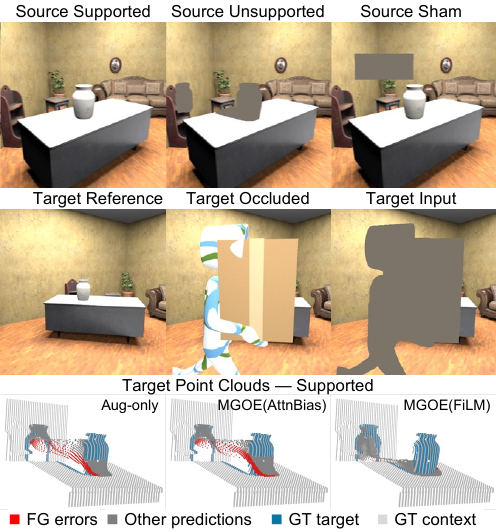}
\caption{\textbf{Cross-view evidence and foreground residuals.}
The clean target reference is not a model input.
Target-view geometry predicted from multi-view inputs is shown from a shared oblique viewpoint, retaining all target-region predictions.
In this example, \armfilm{} leaves fewer foreground residuals than \armaug{} and \armbias{}.}
\label{fig:cross-view-evidence}
\end{figure}

To examine how the models use cross-view observations during object exclusion, we construct a procedural dataset with known background geometry and controllable source-view support.
This diagnostic tests whether \armaug{} already relies on supporting observations in other views and whether FiLM further reduces foreground residuals in masked regions.
Models are compared on identical cases spanning multiple random seeds, textures, viewpoints, and scene geometries; experimental details are provided in Supplementary \cref{sec:supp-synthetic-diagnostic}.
With the masked target held fixed, the source views either retain the hidden surface (supported), mask its supporting observations (unsupported), or mask an equal-area unrelated region (sham).
Target-region AbsRel measures depth accuracy; under supported inputs, the foreground-direction error rate (FG) measures the fraction of target-region pixels satisfying $\widehat D<D-\max(0.10\,\mathrm{m},0.05D)$.
Here, $\widehat D$ is scale-aligned predicted depth and $D$ is ground-truth depth; FG serves as a proxy for foreground residuals.
\Cref{fig:cross-view-evidence} illustrates this intervention and compares target-view geometry from \armaug{}, \armbias{}, and \armfilm{} under supported inputs.
For visualization, predicted depths are scale-aligned using the unmasked control region and backprojected with ground-truth intrinsics.
For complementary qualitative inspection of transfer to real object masks, sensitivity to mild mask dilation, and foreground residuals on DAVIS2016~\cite{perazzi2016benchmark}, see Supplementary \cref{sec:supp-davis-qualitative}.

\begin{table}[ht]
\centering
\caption{\textbf{Supported recovery and foreground residuals.}
Target-region AbsRel is reported for supported (Sup.), unsupported (Unsup.), and sham inputs.
FG: foreground-direction error rate under supported inputs.
All metrics use depth scaled by the median GT-to-predicted depth ratio on the same unmasked control region, fitted independently for each model, case, and input condition.
Entries are means over the same 96 cases.}
\label{tab:cross-view-support}
\small
\setlength{\tabcolsep}{4pt}
\renewcommand{\arraystretch}{1.12}
\begin{tabular}{@{}lrrrr@{}}
\toprule
& \multicolumn{3}{c}{Target-region AbsRel $\downarrow$} & FG (\%) $\downarrow$ \\
\cmidrule(lr){2-4}
Model & Sup. & Unsup. & Sham & \\
\midrule
\armstock{} & 0.8384 & 0.8357 & 0.8305 & 100.00 \\
\armft{} & 0.8087 & 0.8075 & 0.8020 & 100.00 \\
\armaug{} & 0.0836 & 0.3230 & 0.0982 & 8.76 \\
\armfilm{} & 0.0989 & 0.3370 & 0.1185 & 4.62 \\
\armbias{} & 0.0880 & 0.3239 & 0.0972 & 8.14 \\
\bottomrule
\end{tabular}
\end{table}

\begin{table*}[t]
\centering
\caption{\textbf{Clean-input geometry.}
All methods receive unmodified RGB; mask-conditioned variants receive an empty mask.
RE10K, ScanNet, and DTU use ten views; ETH3D uses 24 views.
ETH3D reports point-map Overall error after scene-level Sim(3) alignment.
ETH3D covers all 13 evaluation scenes.}
\label{tab:clean-geometry}
\small
\setlength{\tabcolsep}{4pt}
\begin{tabular}{@{}lrrrrrr@{}}
\toprule
& \multicolumn{2}{c}{RE10K pose AUC $\uparrow$} & \multicolumn{2}{c}{ScanNet depth} & DTU & ETH3D \\
\cmidrule(lr){2-3}\cmidrule(lr){4-5}
Model & $15^\circ$ & $30^\circ$ & AbsRel $\downarrow$ & $\delta<1.25$ $\uparrow$ & Overall (mm) $\downarrow$ & Overall (m) $\downarrow$ \\
\midrule
\armstock{} & 0.5961 & 0.7418 & 0.0265 & 0.9851 & 2.5108 & 0.3207 \\
\armft{} & 0.5838 & 0.7346 & 0.0236 & 0.9891 & 2.5716 & 0.2438 \\
\armaug{} & 0.5895 & 0.7404 & 0.0245 & 0.9879 & 3.1956 & 0.2483 \\
\armbias{} & \textbf{0.6053} & \textbf{0.7512} & \textbf{0.0234} & \textbf{0.9892} & 3.3618 & 0.2627 \\
\armfilm{} & 0.5982 & 0.7462 & 0.0238 & 0.9887 & \textbf{2.3990} & \textbf{0.2390} \\
\bottomrule
\end{tabular}
\end{table*}

\paragraph{Recovery with cross-view support.}
Synthetic occlusion training reduces supported target-region AbsRel by nearly an order of magnitude, from 0.8087 for \armft{} to 0.0836 for \armaug{} (\cref{tab:cross-view-support}).
For \armaug{}, removing source support raises the error to 0.3230, whereas sham masking yields 0.0982.
These results show that \armaug{} already uses cross-view observations to recover the hidden surface before explicit mask conditioning is added.

\paragraph{Foreground residuals.}
FiLM reduces FG from 8.76\% for \armaug{} to 4.62\% for \armfilm{}.
Paired comparisons do not establish a clear AbsRel difference between \armfilm{} and \armaug{} in this diagnostic; the clearer benefit of FiLM is its reduction in FG.

\subsection{Geometry Estimation on Clean Inputs}
\label{sec:clean-geometry}

We evaluate whether object-exclusion training preserves general-purpose geometry on unmodified inputs, covering camera pose, depth, and 3D reconstruction (\cref{tab:clean-geometry}).
\armfilm{} improves the displayed RE10K, DTU, and ETH3D aggregates over \armft{}, with slightly higher ScanNet error; relative to pretrained VGGT, it has better displayed results on all four datasets.
\armbias{} has the highest RE10K AUC and lowest ScanNet error, while \armfilm{} performs better on DTU and ETH3D.
These results show dataset-dependent trade-offs rather than uniform improvement.
Supplementary \cref{sec:supp-clean-protocol} provides sampling, alignment, and aggregation details.

\subsection{Analysis of Mask Conditioning}
\label{sec:mechanism-experiments}

We analyze attention patterns on Bonn and test whether combining FiLM with attention bias further improves object exclusion.
For masked queries, \armfilm{} reduces mask-occupancy-weighted attention to masked keys from 0.1227 for \armaug{} to 0.1060, while increasing attention to visible background in other views from 0.5644 to 0.5759.
\armbias{} remains close to \armaug{} on both statistics, consistent with its small learned suppression coefficients (mean 0.00605).
Combining the mechanisms in \armfilmbias{} shifts these attention statistics further but yields no clear additional reduction in background depth error or foreground leakage over \armfilm{} in paired Bonn comparisons.
Together, these observations are consistent with FiLM improving the treatment of masked features, with attention redistribution as one observable change rather than an established cause of the geometry gains.
The measurement protocol, layer-wise profiles, conditioning coefficients, and paired results are provided in Supplementary \cref{sec:supp-mechanism-analysis}.

\subsection{Auxiliary Appearance Supervision}
\label{sec:auxiliary-supervision}

Recovering a colored point cloud after object exclusion requires restoring appearance as well as geometry in the masked regions.
We explore incorporating appearance recovery into training through an RGB completion head that predicts colors from shared features and GeoPhoto, which supervises predicted geometry through cross-view color reprojection.
Stronger RGB supervision improves the evaluated appearance metrics, but neither auxiliary approach provides a consistent additional geometric benefit (Supplementary \cref{sec:completion-ablation}).
These findings motivate training the main model with geometric supervision and restoring point-cloud colors through the explicit cross-view transfer in \cref{sec:cross-view-fusion}, which leaves predicted geometry unchanged.

\section{Limitations}

When a surface is occluded or excluded in every input view, \method{} lacks direct cross-view observations to constrain its geometry, and its reconstruction may be inaccurate.

Imprecise exclusion masks can leave foreground pixels unmasked, introducing color contamination in the recovered region.
Such residual observations may also affect depth prediction, although we have not quantified this effect.
Connected-component-based mask cleanup, hole filling, and modest dilation could mitigate incomplete coverage, at the cost of potentially masking useful background observations.

\section{Conclusion}

We introduced \method{}, a mask-guided framework for excluding user-specified objects and recovering background geometry from other input views in feed-forward 3D reconstruction.
Our approach combines geometry-guided synthetic occlusion training with explicit mask conditioning, teaching the model to reconstruct masked surfaces while preserving their original geometric supervision.
Bonn ablations show that synthetic occlusion training improves background recovery, while FiLM conditioning further reduces foreground leakage and background depth error.
Controlled interventions further demonstrate that the learned recovery depends on supporting observations in other views.

{\small
\makeatletter\@startbibunitorrelax\makeatother

}
\end{bibunit}

\begin{bibunit}
\clearpage
\setcounter{section}{0}
\setcounter{subsection}{0}
\setcounter{equation}{0}
\setcounter{figure}{0}
\setcounter{table}{0}
\setcounter{footnote}{0}
\setcounter{page}{1}
\renewcommand{\thesection}{S\arabic{section}}
\renewcommand{\theequation}{S\arabic{equation}}
\renewcommand{\thefigure}{S\arabic{figure}}
\renewcommand{\thetable}{S\arabic{table}}
\renewcommand{\thefootnote}{S\arabic{footnote}}
\renewcommand{\thepage}{S\arabic{page}}
\renewcommand{\theHsection}{supp.\arabic{section}}
\renewcommand{\theHequation}{supp.\arabic{equation}}
\renewcommand{\theHfigure}{supp.\arabic{figure}}
\renewcommand{\theHtable}{supp.\arabic{table}}
\maketitlesupplementary

\section{Training and Implementation Details}
\label{sec:supp-training-protocol}

\paragraph{Controlled fine-tuning baseline.}
Our experiments adapt a pretrained VGGT checkpoint to object exclusion using its public fine-tuning framework and a fixed training protocol.\footnote{\url{https://github.com/facebookresearch/vggt/blob/main/training/README.md}}
The controlled fine-tuning baseline, \armft{}, shares the initialization, data mixture, preprocessing, frame sampling, and two-stage training schedule of the main method.
It disables additional synthetic occlusion and mask conditioning, while retaining dataset-required exclusions.
Comparing \armstock{} with \armft{} measures the combined effect of our data, preprocessing, augmentation, and optimization choices.
Our ablations compare \armft{} with \armaug{} to assess synthetic occlusion training, and \armaug{} with \armfilm{} and \armbias{} to assess the two mask-conditioning mechanisms.

\paragraph{Data distribution and preprocessing.}
Our fixed mixture comprises CA-1M~\cite{lazarow2025cubify}, ScanNet++~\cite{yeshwanth2023scannetpp}, Hypersim~\cite{roberts2021hypersim}, BlendedMVS~\cite{yao2020blendedmvs}, KITTI-360~\cite{liao2023kitti360}, and Co3Dv2~\cite{reizenstein2021co3d}, with sampling weights of 30:15:15:15:15:10.
Offline preprocessing constructs geometric targets and provides quality measurements for selecting training samples according to the criteria below.
For Hypersim, we convert ray distances to camera-axis depth and transform cameras to the common coordinate convention.
Frame selection is also dataset-specific: CA-1M and ScanNet++ sample frames around a random anchor with a minimum temporal gap of 0.3\,s and a maximum span of 12\,s, while Co3Dv2 uses camera-pose proximity.
These choices define our adaptation protocol and are shared across the controlled variants.

\paragraph{Indoor data preparation.}
CA-1M uses the full official training capture list, scene-disjoint splits, and scan-rendered depth with matching ground-truth cameras.
For ScanNet++, we undistort registered iPhone frames to $640\times480$, update intrinsics, and ray-cast depth from the laser mesh, excluding anonymized regions, missing hits, and depths outside $[0.05,20]$\,m.
For both datasets, we check rendered depth against device LiDAR measurements in $(0.1,6)$\,m wherever both are valid.
A pixel is inconsistent when the absolute depth difference exceeds the larger of $0.05$\,m and $5\%$ of its LiDAR depth; these measurements are used only for quality checks.
The frame inconsistency rate is the fraction of valid comparison pixels that are inconsistent.
We exclude frames with rates above $30\%$, scenes whose median per-frame inconsistency rate exceeds $30\%$, and samples lacking a valid comparison.

\paragraph{KITTI-360 background supervision.}
Static point clouds provide background depth through z-buffered splatting with radii capped at six pixels.
We select vehicle and person instances whose identifiers occur in the dynamic point cloud or whose image labels agree with the rendered static-map semantic labels at fewer than $50\%$ of their pixels.
Their masks form $B_s$ after three-pixel dilation, while the rendered background depth behind them is retained for supervision.
Outside $B_s$, valid depth requires agreement between the rendered semantic label and the image annotation; missing depth, void, and sky are excluded.
Frames overlapping official validation windows belong to validation, and each sampled sequence stays within one window.

\paragraph{Training schedule and augmentation.}
Our adaptation uses eight first-stage epochs and sixteen second-stage epochs, with 800 optimization steps per epoch (19,200 steps total).
The DINOv2~\cite{oquab2024dinov2} tokenizer remains frozen throughout; the alternating-attention backbone is frozen in the first stage and fine-tuned in the second.
Results use one training run per variant; view-sampling seeds and paired resampling measure evaluation variability, not variability across independently trained models.
Color jitter, grayscale conversion, and Gaussian blur are disabled in our shared training configuration, while random crop-scale augmentation is retained.

\paragraph{Optimization and checkpoint selection.}
All controlled variants use AdamW with weight decay $0.05$ and the same learning-rate schedule, restarted at each training stage.
The base learning rate rises linearly from $10^{-8}$ to $5\times10^{-5}$ over the first 5\% of each stage and then decays to $10^{-8}$ with a cosine schedule.
In stage~2, the alternating-attention backbone and camera/register tokens use $0.1\times$ this rate, while prediction heads and new modules use the base rate.
We clip gradient norms to 1.0 separately for the backbone and head groups.
Stage~2 starts from the final stage~1 checkpoint, and all reported fine-tuned variants use the final checkpoint after the sixteen stage~2 epochs.

\subsection{Hardware and Computational Cost}
\label{sec:supp-compute}

\paragraph{Training cost.}
We report training costs for the main \armfilm{} configuration.
Both stages ran on four NVIDIA L20 GPUs (48\,GB each) on one node, using bf16 mixed precision, 518-pixel inputs, and 2--24 views per sample with at most 24 images per GPU per forward pass and no gradient accumulation.
The two stages took 11.2\,h and 24.8\,h of wall-clock time, respectively, excluding queueing and the interval between stages.
With four GPUs, these durations correspond to 143.8 GPU-hours in total.

\paragraph{Inference cost.}
We compare \armstock{} and \armfilm{} using a fixed random 24-view input at the Bonn evaluation resolution $518\times392$ (width$\times$height), with batch size one and bf16 precision on one NVIDIA A40 GPU (48\,GB) using PyTorch~2.3.1 and CUDA~12.1.
Both models use the same input and the same camera and depth heads; \armfilm{} additionally receives an all-zero mask.
After 10 warm-up passes, we average 50 timed passes, synchronizing CUDA immediately before and after each forward pass.
The measured interval covers the model forward pass only; checkpoint loading, preprocessing, host--device transfer, and post-hoc RGB fusion are excluded.
\Cref{tab:supp-inference-cost} reports the mean forward time per multi-view sample and the peak allocated GPU memory.
Both models take approximately 2.0\,s per sample and use approximately 12.5\,GiB of peak allocated GPU memory in this setting.

\begin{table}[ht]
\centering
\caption{\textbf{Inference cost on a 24-view input.}
Mean model-forward time and peak allocated GPU memory over 50 synchronized passes after 10 warm-up passes.
Both models use batch size one, bf16 precision, the same fixed random 24-view input, and the same camera and depth heads on one NVIDIA A40.
Checkpoint loading, preprocessing, transfers, and RGB fusion are excluded.}
\label{tab:supp-inference-cost}
\small
\setlength{\tabcolsep}{3pt}
\begin{tabular}{@{}lrr@{}}
\toprule
Model & Forward time (s) & Peak allocated memory (GiB) \\
\midrule
\armstock{} & 1.998 & 12.45 \\
\armfilm{} & 2.010 & 12.47 \\
\bottomrule
\end{tabular}
\end{table}

\subsection{Prediction Heads and Training Objective}
\label{sec:supp-heads-losses}

\paragraph{Shared head configuration.}
We enable the camera and depth heads and disable the dedicated point-map and tracking heads throughout fine-tuning, adopting the public fine-tuning configuration unchanged.
Point maps are obtained by back-projecting predicted depth with the predicted camera parameters.
All controlled fine-tuning variants share these geometric heads and the same point-map construction, so head selection is held fixed when comparing synthetic occlusion, mask conditioning, and auxiliary supervision.

\paragraph{Geometric objective.}
We use the camera and depth losses from the public training code for VGGT~\cite{wang2025vggt} and weight them as follows:
\begin{equation}
\mathcal{L}_{\mathrm{geom}}
=5\mathcal{L}_{\mathrm{camera}}+\mathcal{L}_{\mathrm{depth}}.
\label{eq:geometric-training-objective}
\end{equation}
The camera objective applies $\ell_1$ supervision to translation, quaternion rotation, and field of view, with component weights $1$, $1$, and $0.5$, respectively, and discounts earlier refinement stages by a factor of $0.6$ per stage.
The depth objective combines depth regression, confidence-weighted regression with logarithmic confidence regularization, and a multiscale gradient term.
It uses the original geometric validity masks; valid masked pixels are exempted from the 98th-percentile outlier filtering so that masking-induced errors remain supervised.
The main method and geometric baselines use this objective without an appearance-completion term; auxiliary variants add the corresponding losses described in \cref{sec:completion-ablation}.

\section{Geometry-Guided Synthetic Occlusion}
\label{sec:supp-synthetic-occlusion}

\begin{figure*}[t]
\centering
\includegraphics[width=\textwidth]{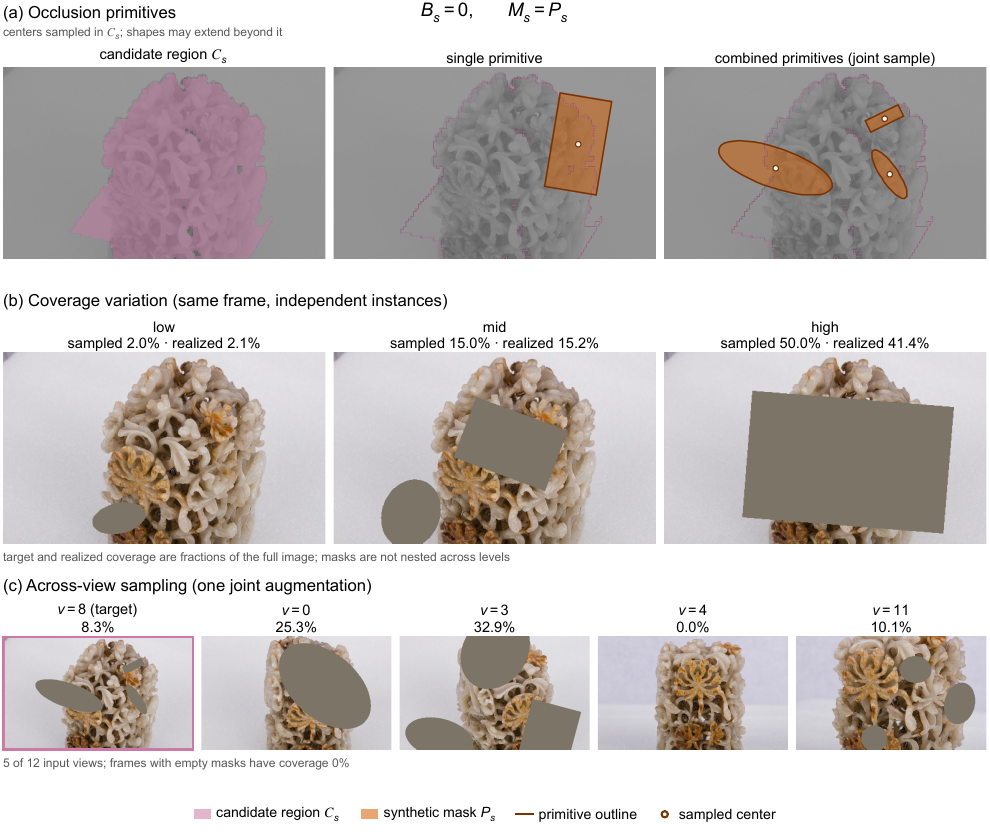}
\caption{\textbf{Geometry-guided occlusion sampling on BlendedMVS.}
(a) The approximate candidate region $\mathcal{C}_s$ and sampled primitives; centers lie within the region, while shapes may extend beyond it.
(b) Independent occlusion samples on the same frame, with target and realized coverage measured relative to the full image area.
(c) Five mean-filled inputs from one twelve-view sample; $v$ is the zero-based input-view index and percentages denote realized coverage.
This sample has $B_s=0$, so $M_s=P_s$.}
\label{fig:supp-synthetic-occlusion-sampling}
\end{figure*}

\begin{figure*}[t]
\centering
\includegraphics[width=\textwidth]{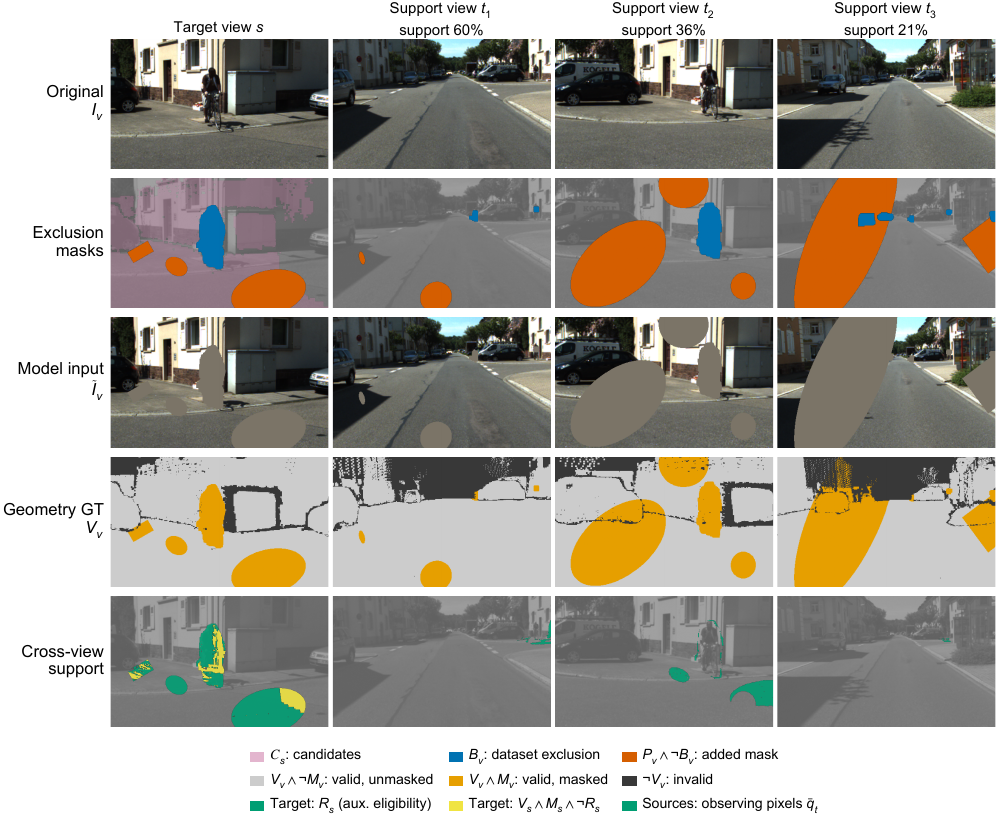}
\caption{\textbf{Dataset exclusions and geometric supervision on KITTI-360.}
Columns show a target $s$ and three support views from a twelve-view sample; $v$ indexes the view in each column.
Pink marks the approximate candidate region, blue marks $B_v$, and orange marks added exclusions $P_v\land\neg B_v$.
The validity row separates valid unmasked, valid masked, and invalid pixels; both valid categories retain geometric supervision.
In the support row, green in the target view marks $R_s$, while yellow marks masked valid pixels without support.
Green in each source view marks pixels providing support under the final masks.
Each source percentage is the fraction of masked valid target pixels supported by that view; a target pixel may be supported by multiple source views.
Across all eleven other views, $18{,}149$ of $23{,}880$ masked valid target pixels retain support (76.0\%); all $23{,}880$ remain eligible for geometric supervision.}
\label{fig:supp-synthetic-occlusion-details}
\end{figure*}

\paragraph{Candidate region construction.}

For each of the $S$ input views indexed by $s$, let $I_s$, $Z_s$, and $V_s$ denote its image, ground-truth depth, and geometric validity mask, and let $B_s$ mark dataset-required exclusions.
The pixel domain is $\Omega_s$, with size $H\times W$.
We use ground-truth camera parameters: intrinsics $K_s\in\mathbb{R}^{3\times3}$ and camera-from-world extrinsics $T_s\in\mathbb{R}^{3\times4}$.

For a fixed target pixel $p\in\Omega_s$, let $\mathbf{X}_{s,p}$ be its ground-truth world point.
Its camera coordinates and image projection in a source view $t\neq s$ are
\begin{equation}
\begin{aligned}
\mathbf{x}_t &= T_t[\mathbf{X}_{s,p}^{\top},1]^{\top},\\
q_t &= \pi(K_t\mathbf{x}_t),
\qquad z_t=(\mathbf{x}_t)_3,
\end{aligned}
\label{eq:supp-candidate-projection}
\end{equation}
where $\pi$ performs perspective division and $z_t$ is the projected depth.
The dependence of $\mathbf{x}_t$, $q_t$, and $z_t$ on the fixed target $(s,p)$ is omitted for readability.
For an in-bounds projection, $\bar q_t$ is the nearest pixel, obtained by rounding and clamping to the image boundary, and $d_t=Z_t(\bar q_t)$ is its ground-truth depth.

Let $U=\{U_t\}_{t=1}^{S}$ specify which source pixels may provide valid support.
We set $\operatorname{Vis}_{s\rightarrow t}(p;U)=1$ when all four conditions below hold, and zero otherwise:
\begin{equation}
\left\{
\begin{aligned}
&z_t>\epsilon,\\
&0\leq(q_t)_x<W,\qquad 0\leq(q_t)_y<H,\\
&U_t(\bar q_t)=1,\\
&|z_t-d_t|\leq\delta\max(d_t,\epsilon).
\end{aligned}
\right.
\label{eq:supp-candidate-visibility}
\end{equation}
Here, $\epsilon=10^{-6}$ is the depth threshold and $\delta=0.05$ is the relative tolerance.
Nearest-neighbor sampling avoids mixing depths across surface boundaries.

For candidate construction, supporting pixels must have valid geometry and lie outside the dataset exclusion mask:
\begin{equation}
U_t^{B}=V_t\land\neg B_t,
\qquad U^{B}=\{U_t^{B}\}_{t=1}^{S}.
\end{equation}
The candidate region is then
\begin{equation}
\mathcal{C}_s=\left\{
 p\in\Omega_s\;\middle|\;
 \begin{aligned}
 &V_s(p)=1,\quad B_s(p)=0,\\
 &\exists t\neq s:\\[-2pt]
 &\operatorname{Vis}_{s\rightarrow t}(p;U^{B})=1
 \end{aligned}
\right\}.
\label{eq:supp-candidate-region}
\end{equation}
Throughout, Boolean operations on mask collections apply pixelwise in each view; for example, $V\land\neg M$ denotes $\{V_t\land\neg M_t\}_{t=1}^{S}$.

For placement, we approximate $\mathcal{C}_s$ on a stride-4 grid, expand positive cells, and intersect them with $V_s\land\neg B_s$; final cross-view eligibility is evaluated at full resolution (\cref{eq:aux-recovery-mask}).

\paragraph{Synthetic occlusion generation.}

Our synthetic occluders use only two primitive shapes: rectangles and ellipses.
For each multi-view sample, we independently select frames for synthetic occlusion.
Given the sequence length $S$, the per-frame selection probability is
\begin{equation}
p_{\mathrm{frame}}
=
\min\left(p_{\mathrm{cap}},\tau^{1/S}\right),
\end{equation}
where $\tau$ is an upper bound on the probability that all frames are selected for synthetic occlusion, and $p_{\mathrm{cap}}$ limits the per-frame probability.
We use $\tau=0.05$ and $p_{\mathrm{cap}}=0.9$.
A selected frame is not guaranteed to retain a nonempty synthetic mask after placement and regeneration.

For each selected frame, the target coverage is sampled from $\{0.02,\allowbreak 0.03,\allowbreak 0.05,\allowbreak 0.08,\allowbreak 0.10,\allowbreak 0.15,\allowbreak 0.20,\allowbreak 0.25,\allowbreak 0.30,\allowbreak 0.40,\allowbreak 0.50\}$.
We subtract the area of $B_s$ from the target masked area to obtain the synthetic occlusion area budget.
We sample one to four primitives with equal probabilities for rectangles and ellipses, aspect ratios in $[0.3,1.0]$, and rotations in $[0^\circ,360^\circ)$; areas follow a symmetric Dirichlet allocation.
Centers are proposed uniformly from the approximate candidate region and accepted subject to inter-primitive separation.
The region constrains centers only: shapes may extend beyond it or overlap $B_s$.
When the candidate region is empty, we propose centers uniformly over an expanded image domain, allowing shapes to extend beyond the image boundary.
An additional primitive may be sampled to compensate for an area shortfall after placement and clipping.
These steps and subsequent mask regeneration can change the primitive count and realized coverage.

Let $P_s$ denote the resulting synthetic occlusion mask.
The final exclusion mask is the union of the existing dataset mask and the generated synthetic mask:
\begin{equation}
M_s = B_s \lor P_s.
\end{equation}
The coverage target applies to this union, and $B_s$ is preserved throughout augmentation.
KITTI-360 is the only training dataset with nonzero dataset exclusion masks; the other five use $M_s=P_s$.
Masked inputs follow \cref{eq:synthetic-masked-input}, replacing excluded RGB with the ImageNet mean.
\Cref{fig:supp-synthetic-occlusion-sampling,fig:supp-synthetic-occlusion-details} illustrate sampling and the distinction between geometric validity and cross-view support.

\paragraph{Mask regeneration.}
All augmentation-based variants use the full-resolution eligibility mask $R_s$ (\cref{eq:aux-recovery-mask}) to encourage cross-view support.
When fewer than 100 masked valid pixels across the input group retain support, we may remove one primitive and regenerate masks on the same selected frames, using at most three generation attempts.
The final masks are retained even below this threshold; regeneration preserves the rectangle-and-ellipse family and does not guarantee support for every masked pixel.

\paragraph{Geometric supervision under masking.}
Original geometry and camera targets remain unchanged, with geometric supervision over $\mathcal{G}_s=\{p\in\Omega_s\mid V_s(p)=1\}$ as in \cref{eq:geometric-supervision-domain}.
Valid masked pixels remain supervised regardless of cross-view support and are exempted from quantile-based depth-outlier filtering.
The auxiliary eligibility mask $R_s$ therefore does not restrict $\mathcal{G}_s$.

\section{Mask Conditioning}
\label{sec:supp-mask-conditioning}

Constant-color filling removes the excluded image content, while the binary mask $M_s\in\{0,1\}^{H\times W}$ explicitly identifies where observations have been withheld.

\subsection{FiLM}
\label{sec:film_conditioning}

Let $\mathbf{F}_s\in\mathbb{R}^{N\times C}$ denote the DINOv2 patch tokens of view $s$, with embedding dimension $C$ and patch size $r$.
For image dimensions divisible by $r$, the number of patches is $N=HW/r^2$.
Two independent convolutional projections map the mask to scale and shift tensors for feature-wise affine modulation (FiLM~\cite{perez2018film}):
\begin{equation}
    \begin{aligned}
        \Delta\boldsymbol{\gamma}_s
        &= \operatorname{Flat}\!\left(
            \operatorname{Conv}_{r,r}^{\gamma}(M_s)\right),\\
        \Delta\boldsymbol{\beta}_s
        &= \operatorname{Flat}\!\left(
            \operatorname{Conv}_{r,r}^{\beta}(M_s)\right).
    \end{aligned}
    \label{eq:mask_projection}
\end{equation}
Each convolution uses kernel size and stride $r$ to map the mask within each image patch to $C$ modulation channels; $\operatorname{Flat}$ arranges the output into the $N\times C$ patch-token layout.
The resulting scale and shift modulate $\mathbf{F}_s$ through \cref{eq:mask_film}, before camera and register tokens are concatenated and before the alternating-attention blocks.
This placement makes the exclusion information available to subsequent query, key, and value computations, allowing the mask to influence both attention allocation and the features exchanged across views.
Both projections are bias-free and zero-initialized, making the FiLM transformation an identity at initialization and, for an all-zero mask, at any learned weights.

\subsection{Attention Bias}
\label{sec:attention_bias_conditioning}

For patch index $k\in\{1,\ldots,N\}$ in view $s$, let $\mathcal{P}_{s,k}$ contain its $r^2$ pixels.
The patch occlusion ratio is
\begin{equation}
    m_{s,k}
    = \frac{1}{r^2}
      \sum_{p\in\mathcal{P}_{s,k}}M_s(p).
    \label{eq:patch_occlusion_ratio}
\end{equation}
Average pooling with kernel size and stride $r$ yields these ratios in $[0,1]$.
Using the masked fraction preserves partial coverage at object boundaries instead of treating every intersected patch as fully excluded.
In \cref{eq:attention_bias}, key token $j$ receives $m_j=m_{s,k}$ for its corresponding patch, or $m_j=0$ for a camera or register token.
The key indices cover one view in frame attention and all concatenated views in global attention.

For each attention type $a$, layer $\ell$, and head $h$, a separate learned parameter $b_{a,\ell,h}$ defines the nonnegative coefficient
\begin{equation}
    \alpha_{a,\ell,h}=\max(0,b_{a,\ell,h}),
    \label{eq:attention_bias_coefficient}
\end{equation}
with $b_{a,\ell,h}$ initialized to zero.
The key-side penalty is applied before softmax in both frame and global attention, without removing masked tokens from the query sequence.
This separates the two roles of a masked position: its key is penalized as a source of information, while its query can still gather observations from other positions for background prediction.
Zero initialization leaves attention unchanged initially, while an all-zero mask eliminates the penalty for any learned coefficients.

\section{Bonn Object-Exclusion Evaluation Protocol}
\label{sec:supp-bonn-protocol}

\paragraph{Inputs and evaluation regions.}
Bonn~\cite{palazzolo2019refusion} provides dynamic RGB-D sequences and static-background reference geometry.
We use the same 20 dynamic videos at all input budgets, with nested frame sets $\mathcal{F}_{10}\subset\mathcal{F}_{16}\subset\mathcal{F}_{24}$ and $|\mathcal{F}_n|=n$.
All budgets are scored on the ten targets in $\mathcal{F}_{10}$.
The recovery region $R_{10}$ contains excluded target pixels with valid background geometry supported by another view in $\mathcal{F}_{10}$.
This region and the distinct foreground evaluation set remain fixed across input budgets.
Thus, additional views change the available observations without changing which pixels are scored.
The sixteen-view visualization in \cref{sec:supp-bonn-qualitative} instead recomputes the recovery region using all sixteen inputs.

\paragraph{Static reference and foreground regions.}
We render Bonn's static laser-scan point cloud into each undistorted $640\times480$ view using the reference camera poses and z-buffering.
Let $d_{\mathrm{obs}}$ and $d_{\mathrm{bg}}$ denote sensor depth and rendered background depth in the sensor depth scale, with tolerance $\tau_{\mathrm{B}}(d)=\max(0.10\,\mathrm{m},0.05d)$.
Invalid projections and isolated depths visible through gaps in the point cloud are discarded.
Where both depths are valid, agreement within $\tau_{\mathrm{B}}(d_{\mathrm{bg}})$ identifies visible static surfaces; pixels with $d_{\mathrm{obs}}>d_{\mathrm{bg}}+\tau_{\mathrm{B}}(d_{\mathrm{bg}})$ are excluded as inconsistent with the static reference.
After this filtering, the usable background reference retains pixels with valid rendered depth that are either visibly static or inside the object-exclusion mask.
Cross-view support additionally requires that the background point project to a visible static, unmasked pixel in another input view with depth agreement under the same tolerance.
Foreground evidence is obtained from $d_{\mathrm{obs}}<d_{\mathrm{bg}}-\tau_{\mathrm{B}}(d_{\mathrm{bg}})$ where both depths are valid, followed by hole filling and removal of connected components smaller than 0.1\% of the image.
The leakage evaluation set intersects this foreground evidence with the selected object masks before dilation and valid background reference depth.

\paragraph{Alignment and metrics.}
For each model, video, input condition, and view count, one depth scale is fitted as the median ground-truth-to-predicted depth ratio over valid, unmasked control pixels pooled across the ten target frames.
The fitted scale is used for both background depth error and foreground leakage; it is re-estimated across input conditions and view counts, even when the control pixels are unchanged.
Background AbsRel is evaluated on $R_{10}$.
Foreground leakage is the fraction of pixels in the foreground evaluation set with valid aligned predictions satisfying $\widehat d<d_{\mathrm{bg}}-\tau_{\mathrm{B}}(d_{\mathrm{bg}})$.
These metrics use distinct pixel sets; leakage alone does not measure accurate background recovery.
Pose ATE is evaluated in meters after an independent Sim(3) fit to the predicted and ground-truth camera centers on the same ten targets; this fit does not use the depth scale.
Tables report medians across videos; differences between displayed medians are descriptive, rather than paired effect estimates.
Bold entries indicate the best median result in each column.

\paragraph{Paired comparisons.}
We compare \armfilm{} with \armaug{} on the same 20 videos, ten target frames, and fixed evaluation regions used in \cref{tab:bonn-exclusion}.
For each input budget and metric, we compute the per-video difference (\armfilm{} minus \armaug{}) and report the median paired difference.
We obtain percentile 95\% confidence intervals from 5,000 video-level bootstrap resamples, without multiple-comparison correction.
\Cref{tab:supp-bonn-fixed10-paired} supports the improvements shown in the main-paper table: \armfilm{} reduces both background AbsRel and foreground leakage at all three input budgets, with all corresponding intervals below zero.

\begin{table}[t]
\centering
\caption{\textbf{Paired object-exclusion results on Bonn.}
Differences are \armfilm{} minus \armaug{} on the main table's common 20-video, fixed-ten-target protocol; negative values favor \armfilm{}.
Leakage differences are expressed as fractions.}
\label{tab:supp-bonn-fixed10-paired}
\small
\setlength{\tabcolsep}{3pt}
\begin{tabular}{@{}llrr@{}}
\toprule
Metric & Views & Median difference & 95\% CI \\
\midrule
AbsRel & 10 & $-0.02530$ & $[-0.07824, -0.00130]$ \\
AbsRel & 16 & $-0.02698$ & $[-0.07917, -0.00160]$ \\
AbsRel & 24 & $-0.02889$ & $[-0.08157, -0.00099]$ \\
\midrule
Leakage & 10 & $-0.23555$ & $[-0.45734, -0.03761]$ \\
Leakage & 16 & $-0.27841$ & $[-0.50940, -0.05649]$ \\
Leakage & 24 & $-0.32891$ & $[-0.46483, -0.05012]$ \\
\bottomrule
\end{tabular}
\end{table}

\paragraph{Mask-filling comparison.}
Box masks are derived from the same object-instance masks and expanded by 4\% on each side; they are not detector predictions.
Pretrained VGGT receives raw RGB or object/box masks filled with mean, black, or white color; all fine-tuned variants use object masks and mean-filled RGB.
Object and box masks leave different unmasked control regions for depth-scale estimation, even when their recovery evaluation pixels match.
Possible explanations for the effect of box filling are discussed in \cref{sec:supp-mask-filling}.

\section{Additional Qualitative Results on Bonn}
\label{sec:supp-bonn-qualitative}

\begin{figure*}[t]
\centering
\includegraphics[width=\textwidth]{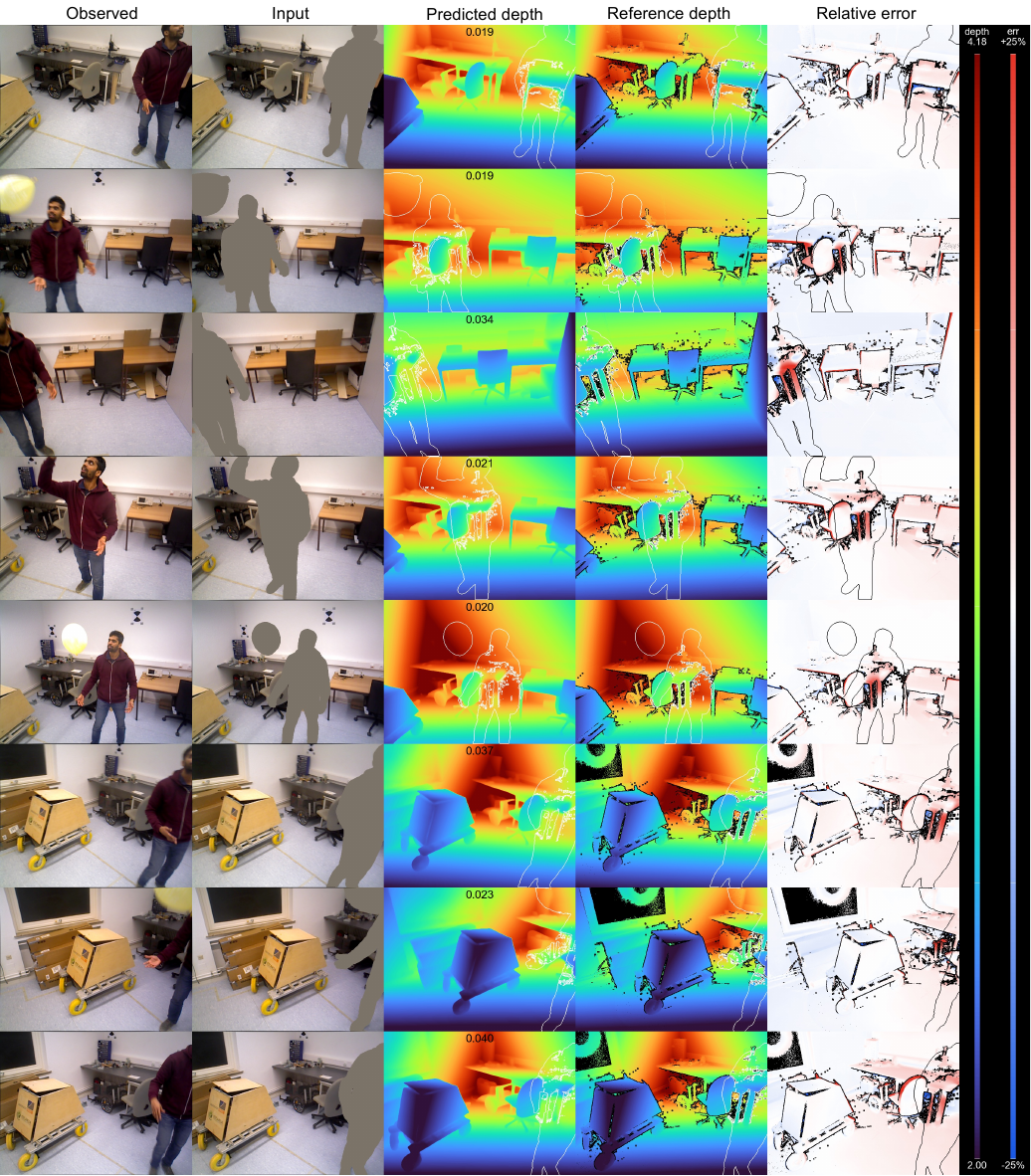}
\caption{\textbf{Multi-view depth visualization in a dynamic Bonn scene.}
Eight of sixteen input views are shown for \armfilm{} with mean-filled object masks.
Numbers report per-frame AbsRel on $R_{16}$, outlined in white on depth and black on error maps.
Relative error $(\widehat D-D_s)/D_s$ is clipped at $\pm25\%$; blue/red indicates nearer/farther predictions.
Black pixels in reference and error maps indicate invalid reference depth.}
\label{fig:supp-bonn-scene}
\end{figure*}

\begin{figure*}[t]
\centering
\includegraphics[width=\textwidth]{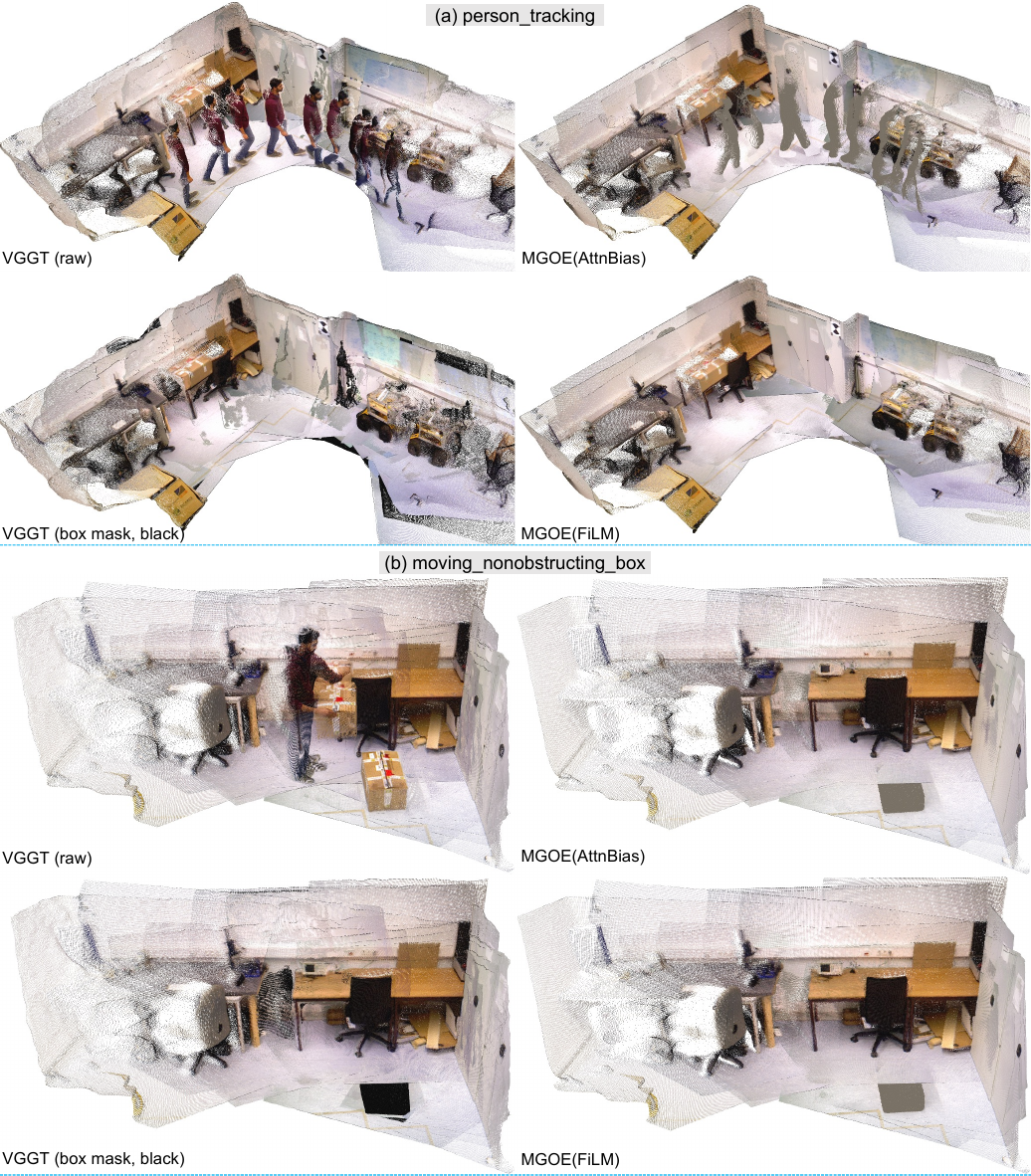}
\caption{\textbf{Point-cloud examples on two dynamic Bonn scenes.}
Groups (a) and (b) show a moving person and a person moving a box, using sixteen input views each.
VGGT uses raw RGB or black-filled box masks; both MGOE variants use mean-filled object masks.
Cross-view color transfer and local color correction leave geometry unchanged.
Gray/black patches retain input fill colors where valid source colors are unavailable.}
\label{fig:supp-bonn-point-clouds}
\end{figure*}

\begin{figure*}[t]
\centering
\includegraphics[width=\textwidth]{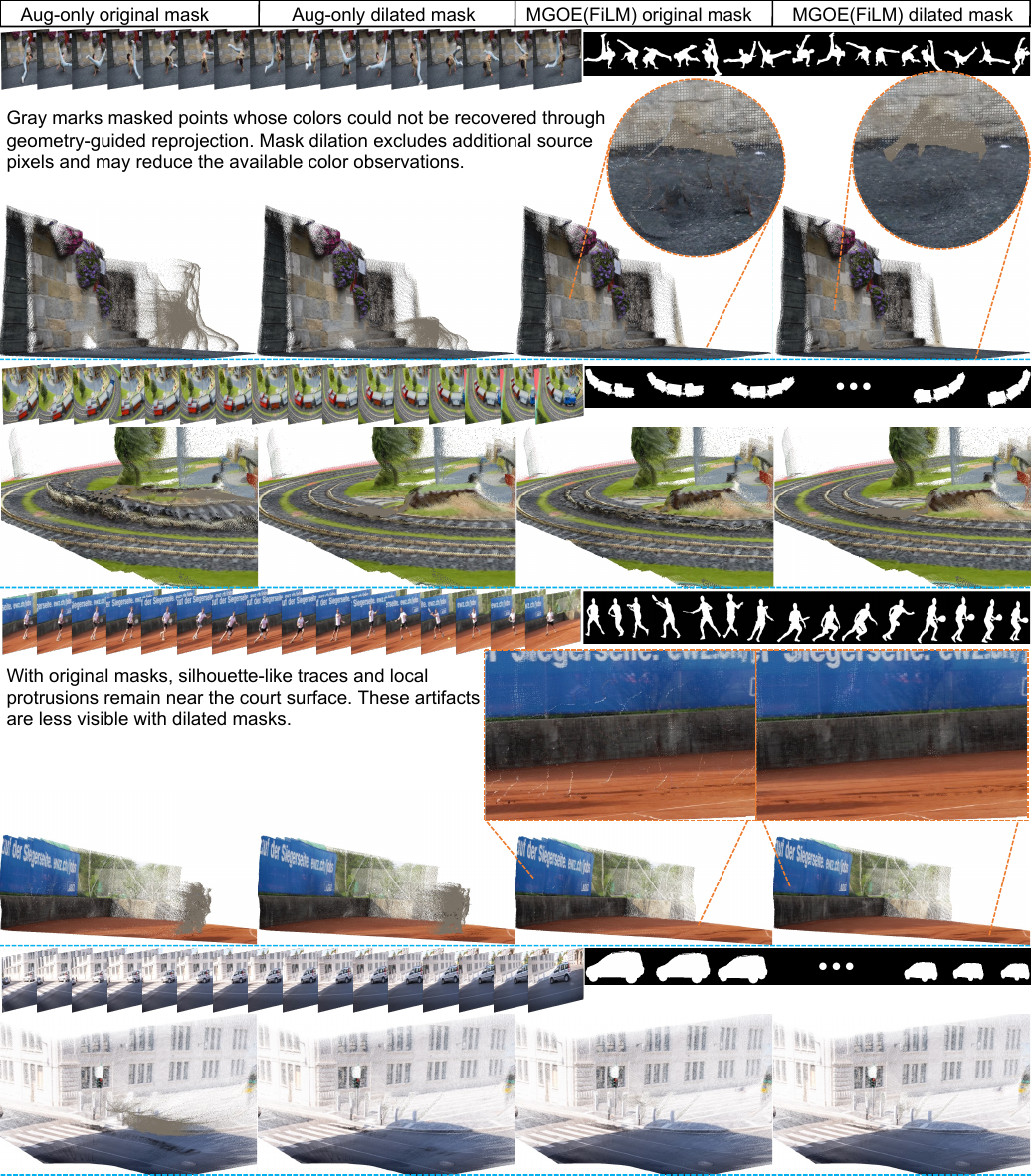}
\caption{\textbf{Qualitative object exclusion on DAVIS2016.}
Columns compare \armaug{} and \armfilm{}, each with original and dilated masks; insets highlight local differences.
Gray masked points retain the input fill color because reprojection found no source color passing the geometric consistency checks.
These points retain their predicted geometry.
Mask dilation reduces ground-surface protrusions across the four illustrated scenes and silhouette traces in tennis, while also excluding potential source pixels for appearance recovery.}
\label{fig:supp-davis-qualitative}
\end{figure*}

\paragraph{Frame-level depth and error.}

\Cref{fig:supp-bonn-scene} shows a person interacting with a balloon (Bonn sequence \texttt{balloon2}), with eight displayed views from a sixteen-view reconstruction by \armfilm{}.
The observed images and mean-filled inputs make the excluded regions explicit, while the predicted depth, static-background reference, and signed error maps show both recovered structure and remaining discrepancies.
Here, $R_{16}$ contains excluded target pixels with valid background geometry supported by at least one other view among the sixteen inputs.
For this visualization, a single depth scale is fitted over valid unmasked control pixels pooled across all sixteen inputs, and per-frame AbsRel is measured on the recovery region $R_{16}$ defined by those sixteen inputs.
Predicted and reference depths share a color scale in Bonn sensor-depth units.
These annotations are therefore distinct from the fixed-$R_{10}$ scores in the main-paper table; the error maps additionally show all pixels with valid reference depth.

\paragraph{Point-cloud comparisons.}
\Cref{fig:supp-bonn-point-clouds} shows a moving person (Bonn sequence \texttt{person\_tracking}) and a person moving a box (\texttt{moving\_nonobstructing\_box}).
All methods use the same sixteen sampled frames per scene: VGGT receives either raw RGB or black-filled box masks, whereas both MGOE variants receive mean-filled object masks and the corresponding conditioning masks.
In the person scene, \armbias{} retains conspicuous person-shaped structures, while \armfilm{} substantially reduces these residuals in the displayed view.
In the box-moving scene, both MGOE variants suppress the prominent foreground objects visible in the raw VGGT reconstruction, but untextured patches remain.
The \method{} reconstructions also show flatter wall surfaces and better visual alignment across views than the VGGT baselines.
This visual improvement is consistent with the lower background depth error and camera trajectory error reported on Bonn.
The qualitative point clouds are displayed without confidence-based filtering.
For visualization, we apply local color correction after the cross-view color transfer described in \cref{sec:cross-view-fusion}, adjusting transferred colors using corresponding visible background pixels just outside the exclusion mask.
This correction reduces brightness and color mismatches between views without changing predicted geometry.
Where no source color passes the consistency checks, points retain the input fill color.

\section{Qualitative Object Exclusion Beyond Bonn}
\label{sec:supp-qualitative-exclusion}

Although the synthetic occluders are composed only of rectangles and ellipses, \armfilm{} also handles irregular object-shaped masks, as demonstrated by the quantitative Bonn results in \cref{tab:bonn-exclusion} and the qualitative Bonn and DAVIS2016 examples in \cref{fig:supp-bonn-point-clouds,fig:supp-davis-qualitative}.

\subsection{DAVIS2016}
\label{sec:supp-davis-qualitative}

We compare \armaug{} and \armfilm{} on DAVIS2016~\cite{perazzi2016benchmark} using the dataset's object annotations and masks dilated by a five-pixel radius at the original 480p resolution.
Each sequence uses the same sixteen sampled frames across both models and mask conditions.
Both models receive mean-filled RGB, and \armfilm{} additionally receives the corresponding exclusion masks.
\Cref{fig:supp-davis-qualitative} shows breakdance-flare, train, tennis, and car-shadow.
Point-cloud colors use post-hoc cross-view appearance recovery (\cref{sec:cross-view-fusion}), which leaves predicted geometry unchanged.

\armfilm{} leaves less prominent foreground-shaped geometry than \armaug{} in the illustrated examples.
Across all four illustrated scenes, mask dilation reduces visible protrusions near the ground surface.
In the tennis example, it also reduces silhouette-like traces near the court surface.
Dilation excludes additional boundary pixels from both the target inputs and potential source observations, illustrating a balance between foreground exclusion and the availability of colors for cross-view transfer.
Gray regions retain the fill color when no unmasked source observation passes the geometric consistency checks.
These examples provide qualitative evidence of mask-boundary sensitivity, without background-depth ground truth for scoring geometric recovery.
The tennis point clouds are not reliably aligned across models, so comparisons focus on local residual shapes rather than pointwise geometric differences.

\subsection{Self-Captured iPhone Scenes}
\label{sec:supp-iphone-qualitative}

We capture eight indoor and outdoor scenes with an iPhone 15 to examine static object exclusion with \armfilm{} using three or four input views.
The bicycle and safety-barrier cases use three views each, and the remaining cases use four.
Exclusion masks are generated with SAM3~\cite{carion2026sam3}, followed by morphological closing and dilation, each with a four-pixel-radius elliptical kernel at the model input resolution.
Point-cloud colors use post-hoc cross-view transfer with local color correction, as in the Bonn visualizations, without changing predicted geometry.

\begin{figure*}[t]
    \centering
    \includegraphics[width=\textwidth]{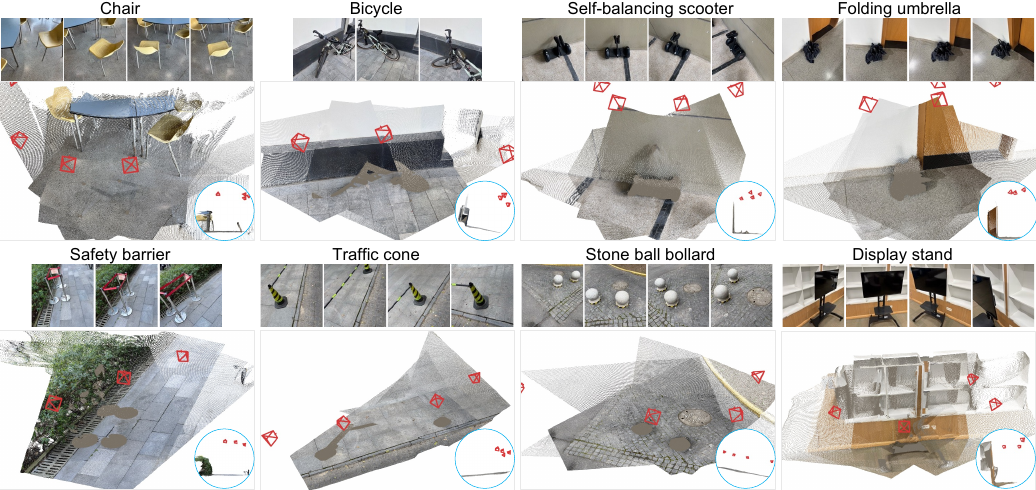}
    \caption{\textbf{Static object exclusion in self-captured iPhone scenes.}
    Each case shows three or four input images above the point cloud reconstructed by \armfilm{}.
    Circular insets at the lower right of each point-cloud panel provide side views of the excluded-object regions, making residual foreground geometry and deviations from the surrounding ground plane easier to inspect.
    Point clouds are colored using post-hoc cross-view transfer with local color correction.
    Gray regions retain the input fill color where no unmasked source observation passes the geometric consistency checks.}
    \label{fig:supp-iphone-static-exclusion}
\end{figure*}

\Cref{fig:supp-iphone-static-exclusion} illustrates static object exclusion across these eight scenes.
The side views show approximately planar ground in several excluded-object regions, with little visible foreground geometry protruding above the surrounding surface.
In the display-stand case, the display occludes a large background area, and the reconstructed surfaces around the shelving remain visibly distorted.

\section{Camera Pose Evaluation}
\label{sec:supp-external-pose}

\paragraph{Evaluation setting and sampled views.}
We evaluate camera pose estimation by dynamic-scene reconstruction methods under the sampled multi-view input setting used for \method{}.
The comparison in \cref{tab:external-pose} uses the same 24-frame samples for every method: 20 dynamic Bonn videos and 26 non-overlapping motion windows from seven TUM RGB-D~\cite{sturm2012benchmark} source sequences.
Bonn inputs place the reference frame first and order the remaining frames chronologically; this produces a backward time step between the first two inputs in six of the twenty samples.
TUM inputs follow chronological order.
All methods receive these sampled views in the same order, with their respective image resizing and normalization.

\paragraph{Method configurations.}
MoRe-FA uses the released full-attention configuration and checkpoint with its native camera head, without global optimization~\cite{fang2026more}.
Easi3R-DUSt3R uses the DUSt3R~\cite{wang2024dust3r} ViT-L backbone, attention-derived masks, and the official flow-based global alignment, with SAM2~\cite{ravi2024sam2} refinement disabled~\cite{chen2025easi3r}.
MonST3R-GO uses the released MonST3R checkpoint, pairwise inference, and official global alignment with SAM2~\cite{ravi2024sam2} refinement enabled~\cite{zhang2025monst3r}.
Both optimization-based methods use the \texttt{swinstride-5-noncyclic} pair graph, minimum-spanning-tree initialization, and 300 optimization iterations with a linear learning-rate schedule starting at 0.01.
Their temporal-smoothing weights are 0 for Easi3R and 0.01 for MonST3R; both retain a flow-loss weight of 0.01.
MoRe uses image size 518, whereas Easi3R and MonST3R use 512 with their respective preprocessing routines.
No ground-truth camera poses, depth maps, or externally supplied exclusion masks are passed to the external methods.
Their internal motion masks and attention are estimated from the input images.

\paragraph{Raw and exclusion-mask conditions.}
In the raw condition, \armstock{} and \armfilm{} receive unmodified RGB, and \armfilm{} receives an all-zero conditioning mask.
The exclusion-mask condition uses the same \armfilm{} checkpoint and frames, fills excluded RGB pixels with the ImageNet mean, and supplies the corresponding mask to the model.
Bonn masks are generated through SAM3~\cite{carion2026sam3} video tracking and sequence-level object selection, covering people, balloons, and selected boxes, with five-pixel dilation at $640\times480$ resolution.
During offline annotation, discrepancies between sensor depth and the rendered static reference help identify foreground objects and reject spurious SAM3 tracks; the selected tracks are manually checked to produce the object masks.
This annotation procedure specifies the objects to exclude, allowing us to evaluate reconstruction given user-specified exclusion masks.
These masks are fixed across the controlled comparisons; reconstruction uses the filled RGB images and, for mask-conditioned models, the binary masks.
Reference depth and camera poses are used to construct and score the evaluation regions, not as reconstruction inputs.
For TUM, we use SAM3~\cite{carion2026sam3} person tracks with manual confirmation to construct exclusion masks, without dilation.
Masks are resized with nearest-neighbor interpolation for model input.
The raw-to-masked comparison measures the joint effect of RGB filling and mask conditioning.

\paragraph{Scoring and aggregation.}
We apply the same evaluator to all methods and independently fit a Sim(3) transform to predicted and ground-truth camera centers for each method, sample, and input condition.
This scoring alignment is separate from any global optimization within a method.
ATE and relative pose error (RPE) follow their standard definitions~\cite{sturm2012benchmark}.
ATE is evaluated over all 24 input frames, unlike the fixed ten target frames in \cref{tab:bonn-exclusion}; relative pose errors use adjacent frames in the actual input order.
Bonn results are medians across the 20 videos.
For TUM, we first take the median over windows within each source sequence, then the median across the seven sequences.
Static reference sequences are excluded from these summaries.
\Cref{tab:supp-external-pose} reports the full 24-view ATE and relative pose errors for both datasets.

\begin{table}[ht]
\centering
\caption{\textbf{Full 24-view pose metrics.}
ATE and translational RPE ($\mathrm{RPE}_t$) are in meters; rotational RPE ($\mathrm{RPE}_r$) is in degrees.
All metrics are lower-is-better.
Raw follows the unmodified-RGB condition in \cref{tab:external-pose}.
Object-mask (Bonn) and person-mask (TUM) inputs use additional exclusion annotations as described above.}
\label{tab:supp-external-pose}
\small
\setlength{\tabcolsep}{3pt}
\begin{tabular}{llrrr}
\toprule
Method & Input & ATE & $\mathrm{RPE}_t$ & $\mathrm{RPE}_r$ \\
\midrule
\multicolumn{5}{l}{Bonn: 20 dynamic videos} \\
MoRe-FA & Raw & 0.0552 & 0.0444 & 1.225 \\
Easi3R-DUSt3R & Raw & 0.0701 & 0.0543 & 2.349 \\
MonST3R-GO & Raw & 0.1154 & 0.0675 & 3.243 \\
\armstock{} & Raw & 0.0394 & 0.0328 & 0.710 \\
\armfilm{} & Raw & 0.0298 & 0.0271 & 0.703 \\
\armfilm{} & Object mask & 0.0264 & 0.0253 & 0.679 \\
\midrule
\multicolumn{5}{l}{TUM: 26 motion windows, seven source sequences} \\
MoRe-FA & Raw & 0.0155 & 0.0120 & 0.755 \\
Easi3R-DUSt3R & Raw & 0.0354 & 0.0191 & 0.889 \\
MonST3R-GO & Raw & 0.0958 & 0.0302 & 6.310 \\
\armstock{} & Raw & 0.0115 & 0.0127 & 0.412 \\
\armfilm{} & Raw & 0.0126 & 0.0132 & 0.384 \\
\armfilm{} & Person mask & 0.0142 & 0.0122 & 0.365 \\
\bottomrule
\end{tabular}
\end{table}

\paragraph{Pose estimation from sampled views.}
Under raw RGB input, \armfilm{} achieves lower median ATE than the three evaluated dynamic-scene reconstruction methods on both Bonn and TUM (\cref{tab:supp-external-pose}).
These results show that a model adapted for static-background reconstruction remains competitive in camera pose estimation from sampled views of dynamic scenes.

\paragraph{Effect of exclusion-mask input.}
We compute paired differences as masked-input error minus raw-input error.
On Bonn, the median paired difference in ATE across 20 videos is $-0.00018$\,m (95\% CI: $[-0.00348,0.00131]$\,m).
On TUM, we compare the 26 windows individually, giving each window equal weight; the table instead aggregates sequences equally.
The median paired difference in ATE is $+0.00049$\,m (95\% CI: $[-5.02\times10^{-9},1.65\times10^{-3}]$\,m).
These percentile intervals use 5,000 bootstrap resamples, resampling videos on Bonn and source sequences on TUM while retaining all windows within each sampled sequence.
The ATE intervals and those for both relative pose metrics include zero on both datasets, leaving the direction of the pose effect unresolved under these input conditions.
The raw rows compare methods without externally supplied masks; the additional rows report \armfilm{} with the dataset-specific exclusion inputs.

\section{Controlled Synthetic Occlusion Diagnostic}
\label{sec:supp-synthetic-diagnostic}

\begin{figure}[ht]
    \centering
    \includegraphics[width=\columnwidth]{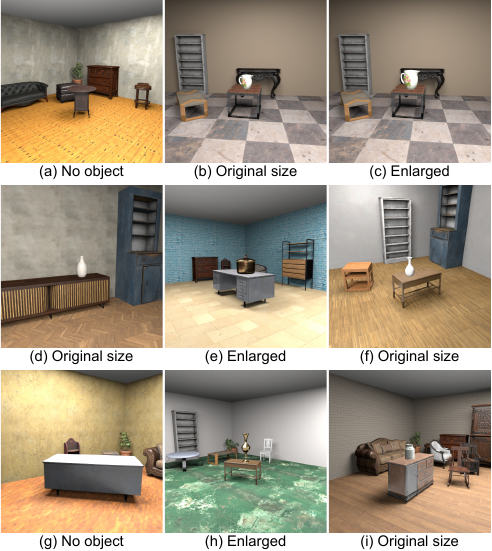}
    \caption{\textbf{Scenes in the controlled synthetic diagnostic.}
    Nine clean target views illustrate the room layouts and retained-object variants.
    Panels (b) and (c) share the same layout, textures, and viewpoint, with the object enlarged by $1.35\times$ in (c).
    Objects in (d) and (h) use mean-color materials.
    These references omit the foreground occluder to reveal the scene being recovered; the masked evaluation inputs are illustrated in \cref{fig:cross-view-evidence}.
    Quantitative results use all 96 cases.}
    \label{fig:supp-synthetic-scenes}
\end{figure}

\paragraph{Purpose and setup.}
The diagnostic in \cref{sec:cross-view-recovery} examines supported-surface recovery after synthetic occlusion training and foreground residuals after adding mask conditioning.
We procedurally generate indoor scenes using eight random seeds, two texture settings, two viewpoint settings, and three scene geometries, yielding 96 cases.
Only target view 0 is scored in each case.
The three geometries contain no target scene object, an object at its original size, or the same object enlarged by a factor of $1.35$; the room and furniture remain in all three.
These scene objects are to be retained behind the foreground occluder.
The two texture settings use the object's original textures or replace them with each material's mean color.
\Cref{fig:supp-synthetic-scenes} shows selected clean target views covering all eight layouts, both texture settings, and all three geometries.
We compare \armft{} with \armaug{} for supported recovery, and \armaug{} with \armfilm{} and \armbias{} for foreground residuals.

\paragraph{Input conditions.}
All models receive the same mean-filled RGB inputs; mask-conditioned models additionally receive the exclusion masks.
The target exclusion mask covers the moving person and box together with the target image region used to test hidden-surface recovery.
Supported, unsupported, and sham share the masked target, camera poses, ground truth, and evaluation regions.
Supported inputs retain source observations of the hidden target surface; unsupported inputs mask these observations, while sham inputs mask an equal-area unrelated region (\cref{fig:cross-view-evidence}).
This intervention uses image-space masking to control source evidence within the procedural scene family.

\paragraph{Evaluation.}
Let $D$ be target ground-truth depth and $\widehat D=sD_{\mathrm{pred}}$, with scale $s$ fitted separately for each model, case, and condition as the median ground-truth-to-predicted depth ratio on the target view's unmasked control region.
The control region is fixed across supported, unsupported, and sham inputs, but each condition receives its own fitted scale.
Both AbsRel and foreground-direction errors use this scaled depth; qualitative point maps use the same alignment rule.
We construct the evaluation region $T$ from the union of the target-view projections of the original and enlarged retained object.
Within this union, we retain only pixels whose surfaces have observations in other views under both supported and sham inputs in all three geometry variants.
Thus, the same target pixels are evaluated when the retained object is absent, present, or enlarged.
With target exclusion mask $M$, the evaluation set is $E=T\cap M\cap\operatorname{valid}(D)$, which equals $T$ in all 96 cases.
Target-region AbsRel measures supported recovery; unsupported and sham provide controls for the use of source observations.
To measure foreground residuals under supported inputs, we use
\begin{equation}
e_{\mathrm{fg}}=\frac{|\{p\in E:\widehat D_p<D_p-\tau(D_p)\}|}{|E|},
\label{eq:synthetic-direction-errors}
\end{equation}
where $\tau(D)=\max(0.10\,\mathrm{m},0.05D)$.
This measures erroneous geometry in front of the desired surface and serves as a proxy for foreground residuals.
Reported metrics are averaged equally over all 96 cases.
Paired bootstrap intervals resample the eight scene seeds as clusters (2,000 resamples).
For \armfilm{} minus \armaug{} under supported inputs, the 95\% interval spans zero for the AbsRel difference ($[-0.0094,0.0398]$), but remains below zero for the foreground-direction error-rate difference ($[-0.0596,-0.0240]$).

\section{Clean-Input Geometry Evaluation Protocol}
\label{sec:supp-clean-protocol}

All methods receive unmodified RGB; mask-conditioned variants receive an all-zero mask.
RE10K, ScanNet, and DTU use ten views, while ETH3D uses 24.
RE10K~\cite{zhou2018stereo} reports camera-pose AUC at $15^\circ$ and $30^\circ$; ScanNet~\cite{dai2017scannet} reports depth AbsRel and $\delta<1.25$.
Following PoseDiffusion~\cite{wang2023posediffusion}, we compute pose AUC up to $\tau$ from the accuracy curve over thresholds $t\in[0,\tau]$, where accuracy at $t$ is the fraction of image pairs with both relative rotation error and translation-direction error at most $t$.
We evaluate these angular errors directly, without fitting a depth scale or aligning the trajectories.
ScanNet uses a fixed set of 792 clips from 100 scenes.
For each ScanNet clip, the predicted target depth is resized to the ground-truth resolution and multiplied by $\operatorname{median}(D)/\operatorname{median}(D_{\mathrm{pred}})$ over valid target pixels.
AbsRel and threshold accuracy follow the standard definitions of Eigen \etal~\cite{eigen2014depth}, and the per-image median alignment follows the robust multi-view depth benchmark~\cite{schroppel2022benchmark}.
This per-image scale is shared by AbsRel and $\delta<1.25$; it is a ratio of medians, rather than the median of pixelwise ratios used for Bonn and the synthetic diagnostic.
DTU~\cite{jensen2014large} reports point-cloud Overall error in millimeters, averaged over three view-sampling seeds.
For each model, scan, and view sample, a Sim(3) transform is fitted to predicted and ground-truth camera centers and applied to the entire predicted point cloud.

ETH3D~\cite{schops2017multi} uses all 13 scenes in the high-resolution multiview training split solely for evaluation in our fine-tuning protocol.
It reports point-map Overall error in meters, rather than official MVS F1.
For each model, scene, and view sample, one Sim(3) transform is fitted using valid corresponding predicted and ground-truth points across all selected views and applied to the entire predicted point cloud.
For each scene, we average results over distinct sampled view sets, then average across scenes.
Bold values in \cref{tab:clean-geometry} indicate the best result in each column.

\section{Analysis of Mask Conditioning}
\label{sec:supp-mechanism-analysis}

\paragraph{Protocol.}
We compare \armaug{}, \armfilm{}, \armbias{}, and their combined FiLM-and-bias variant, \armfilmbias{}.
All four models receive the same ten-view inputs and mean-filled object masks for 22 Bonn dynamic videos.
The ten-view setting provides 22 evaluable videos; \cref{tab:bonn-exclusion} uses the 20 videos available at all three input budgets.
Paired results for the main table's 20-video fixed-target protocol are reported in \cref{tab:supp-bonn-fixed10-paired}; here, geometry and attention are analyzed on the same 22-video diagnostic set.
Depth error and leakage use the control-region median-ratio scaling in \cref{sec:supp-bonn-protocol}, fitted independently for each model and video over its ten input frames.
Attention statistics use fixed query subsamples and are averaged across videos, global-attention layers, and heads.
Masked-key mass weights attention by each key patch's mask occupancy; cross-visible mass measures attention assigned to visible background keys in other input views.
A source patch is counted as visible background when more than half of its pixels are both unmasked and classified as visible static surfaces by the reference geometry (\cref{sec:supp-bonn-protocol}).
This statistic sums attention to those patches in other views without requiring correspondence to the query surface.

\begin{table}[t]
\centering
\caption{\textbf{Global attention from masked queries.}
Both columns report mean attention mass under the ten-view diagnostic protocol.}
\label{tab:supp-mechanism-attention}
\small
\setlength{\tabcolsep}{3pt}
\begin{tabular}{@{}lrr@{}}
\toprule
Model & Masked-key mass & Cross-visible mass \\
\midrule
\armaug{} & 0.1227 & 0.5644 \\
\armfilm{} & 0.1060 & 0.5759 \\
\armbias{} & 0.1228 & 0.5651 \\
\armfilmbias{} & 0.1023 & 0.5781 \\
\bottomrule
\end{tabular}
\end{table}

\begin{figure*}[t]
\centering
\includegraphics[width=\textwidth]{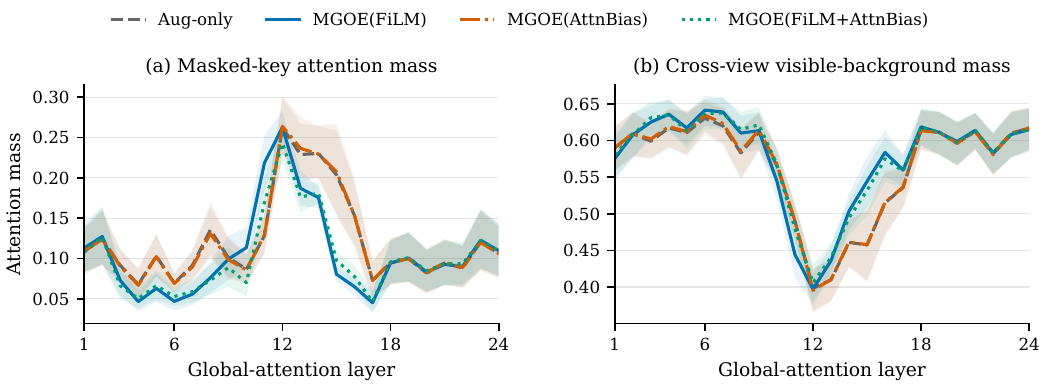}
\caption{\textbf{Layer-wise attention allocation from masked queries.}
(a) Mask-occupancy-weighted attention mass assigned to masked keys; (b) attention mass assigned to visible background keys in other input views.
At each global-attention layer, we average over heads within each video and then equally over the 22 Bonn videos.
Shading denotes pointwise 95\% percentile bootstrap confidence intervals from 10,000 video resamples, reflecting evaluation-video variability rather than training-seed variability.
Layer indices are one-based; all models use the same ten-view inputs and query-sampling protocol.}
\label{fig:supp-attention-layers}
\end{figure*}

\paragraph{Learned coefficients and attention patterns.}
The effective nonnegative attention-bias coefficients in \armbias{} have mean 0.00605 and maximum 0.0727, with 21.6\% of layer--head coefficients exceeding $10^{-3}$.
In \armfilmbias{}, the mean falls to 0.00051 and only 3.8\% exceed this threshold.
Most bias coefficients are small, and the bias-only model assigns nearly the same aggregate attention mass to masked keys as \armaug{} (\cref{tab:supp-mechanism-attention}).
FiLM modulates token representations before alternating attention, allowing it to affect subsequent queries, keys, and values; the explicit attention-bias operation penalizes masked-key logits.
Compared with \armaug{}, \armfilm{} has lower masked-key mass and higher cross-visible mass.
\Cref{fig:supp-attention-layers} shows that this redistribution varies across layers, with a pronounced separation around layers 13--17, while the bias-only curves closely track \armaug{}.

\paragraph{Paired geometry comparisons.}
We report median paired differences across the 22 videos under the masked-input condition, with percentile 95\% confidence intervals from 5,000 video-level bootstrap resamples.
For \armfilm{} minus \armaug{}, the median paired differences are $-0.0328$ in recovery-region AbsRel (95\% CI: $[-0.0808,-0.0021]$) and $-0.310$ in foreground leakage ($[-0.495,-0.077]$).
For \armbias{} minus \armaug{}, the corresponding differences are $-0.00063$ ($[-0.0033,0.0147]$) and $0.0108$ ($[-0.0077,0.0547]$).
\armfilmbias{} yields differences of $-0.00058$ ($[-0.0016,0.0004]$) in AbsRel and $-0.00012$ ($[-0.0028,0.0021]$) in leakage relative to \armfilm{}.
The combined model therefore shows no clear additional recovery benefit in this diagnostic.
The intervals characterize variation across evaluation videos for these checkpoints.

\paragraph{Interpretation.}
In the synthetic diagnostic, \armaug{} recovers hidden surfaces more accurately when source observations are available.
On Bonn, \armfilm{} improves background recovery and reduces foreground leakage relative to \armaug{}, while assigning less attention to masked keys and more to visible background keys in other views.
A plausible interpretation is that feature modulation helps the model treat masked inputs as missing observations, affecting both information selection and the features passed downstream.
Adding attention bias to FiLM further shifts attention toward visible source keys but yields no clear additional geometric benefit in the paired comparisons.

\section{Discussion of Mask-Filling Baselines}
\label{sec:supp-mask-filling}

\paragraph{Object-shaped versus rectangular filling.}
Filling a tight object mask removes its texture but preserves an object-shaped boundary in the input.
A box mask replaces that boundary with a rectangle and also removes surrounding pixels, including potentially useful background observations.
These changes may alter the cues used by pretrained VGGT, providing a possible explanation for the lower background error and foreground leakage observed with box filling.
This comparison measures the combined effect of boundary shape, excluded area, and the control region used for depth-scale fitting.
The color comparisons further show that lower background depth error need not coincide with lower pose error.

\paragraph{Relation to occlusion robustness and pretraining.}
Bratuli\'{c} \etal~\cite{bratulic2026geometric} report that VGGT retains some correspondence-matching ability under white-patch occlusion, with only a small increase in Sampson error in their controlled experiment.
They suggest a possible connection to DINOv2's iBOT pretraining objective, which predicts teacher targets at masked patch positions~\cite{oquab2024dinov2,zhou2022ibot}.
This objective operates on patch tokens, whereas our filling baseline replaces RGB pixels with a constant color at inference.

\section{Auxiliary Appearance Supervision}
\label{sec:completion-ablation}

\begin{figure*}[t]
    \centering
    \includegraphics[width=\textwidth]{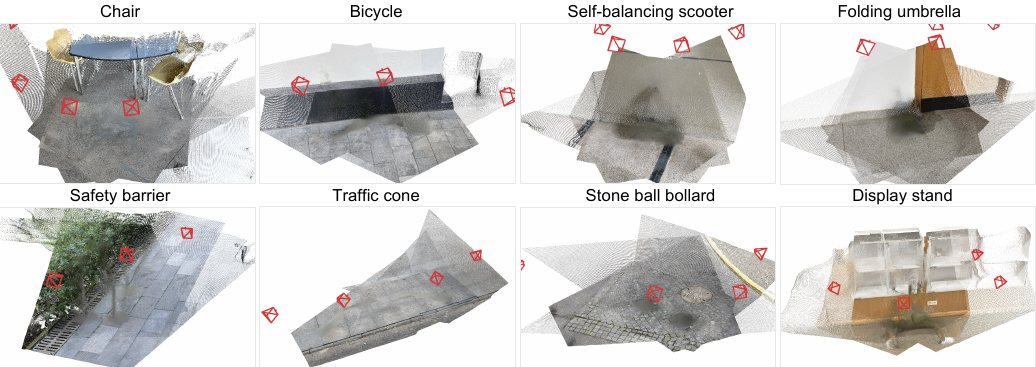}
    \caption{\textbf{RGB completion on self-captured iPhone scenes.}
    Results use \armrgb{} with $\lambda=1.5$ and correspond case by case to \cref{fig:supp-iphone-static-exclusion}, using the same input views and exclusion masks.
    Within masked regions, point-cloud colors come from cross-view transfer with local color correction where a valid source is available, and from the RGB completion head elsewhere.
    Predicted colors replace the constant fill in regions without valid transferred colors, although blurred textures and color mismatches remain visible.}
    \label{fig:supp-rgb-completion-iphone}
\end{figure*}

\paragraph{Motivation and relation to color transfer.}
Cross-view color transfer restores appearance where a source observation passes the geometric consistency checks, leaving the gray input fill in the remaining masked regions.
This motivates two questions: whether shared reconstruction features can support learned appearance completion, including where color transfer fails, and whether appearance supervision also improves background geometry and foreground exclusion.
The RGB variant addresses the first question with a completion head that predicts colors from the shared features and trains the backbone through RGB regression.
To examine the geometric effects of appearance supervision, we compare this feature-based prediction with GeoPhoto, which reprojects source colors using predicted depth and cameras and applies a photometric loss to those geometric outputs.
GeoPhoto relies on source observations and introduces no learnable RGB head.
Both variants retain the camera and depth prediction heads and geometric objective in \cref{sec:supp-heads-losses}, adding their respective auxiliary appearance losses.
They are distinct from the post-hoc fusion in \cref{sec:cross-view-fusion}, which transfers observed colors using fixed model predictions and updates neither the network nor the predicted geometry.

\paragraph{Cross-view eligibility for auxiliary supervision.}
The binary mask $R_s$ identifies occluded, geometrically valid pixels that retain cross-view observation support in the final model input.
Using the visibility test in \cref{eq:supp-candidate-visibility}, we define
\begin{equation}
\begin{aligned}
R_s(p)
={}&M_s(p)\land V_s(p)\\
&\land
\mathbb{1}\left[
\exists t\neq s:
\operatorname{Vis}_{s\rightarrow t}
(p;V\land\neg M)=1
\right].
\end{aligned}
\label{eq:aux-recovery-mask}
\end{equation}
The support mask $V_t\land\neg M_t$ requires the projected surface to remain geometrically valid and unmasked in another actual input view.
We evaluate visibility at full resolution using the original camera and depth coordinates and nearest-neighbor sampling.
This visibility test defines cross-view eligibility for auxiliary supervision and does not restrict the original geometric supervision domain $\mathcal{G}_s$.
The training mask $R_s$ is distinct from the Bonn evaluation region $R_{10}$ in \cref{tab:bonn-exclusion}.

\paragraph{RGB completion supervision.}
Let $Q_s\in\{0,1\}^{H\times W}$ denote RGB-target validity, indicating whether the original RGB value depicts the surface represented by the geometry target.
For example, foreground RGB over a dataset-provided background depth target is not a valid completion target.
The auxiliary completion head is supervised on
\begin{equation}
\mathcal{A}=\{(s,p)\mid R_s(p)=1,\;Q_s(p)=1\}.
\label{eq:completion-supervision-domain}
\end{equation}
On this domain, the objective averages the unsquared Euclidean RGB error between the predicted color and the original image $I_s(p)$, without confidence weighting.
We omit the completion loss when fewer than 100 pixels across the batch belong to $\mathcal{A}$.
\paragraph{Geometry-driven photometric supervision (GeoPhoto).}
GeoPhoto uses the same target domain $\mathcal{A}$, with supporting source views selected in advance from ground-truth geometry and source RGB validity.
This eligibility check requires all four bilinear source neighbors to be geometrically valid and unmasked.
As in differentiable view synthesis~\cite{zhou2017unsupervised}, we use predicted depth and camera parameters to sample colors from the selected source views and compare them with the observed target RGB.
Errors are averaged over eligible sources, with gradients flowing through the projection to predicted depth and cameras.
The RGB completion head, by contrast, predicts color directly rather than through an explicit reprojection operator.
Neither auxiliary objective is used in the main \armfilm{} configuration.

\paragraph{Ablation results.}

We separately evaluate RGB completion (\armrgb{}) and geometry-driven photometric supervision (\armgeo{} and \armbiasgeo{}).
The suffixes RGB and GeoPhoto denote auxiliary appearance regression and geometry-based photometric supervision, respectively; loss weights are reported separately.
Appearance metrics evaluate the RGB head output before cross-view color fusion on fixed, synthetically occluded validation samples from the six training datasets, with equal weight per dataset.
PSNR uses pixel errors in the cross-view-supported masked region; SSIM and spatial LPIPS are averaged over the same region, using the original images as reference.
Increasing the RGB loss weight from 0.05 to 1.5 improves recovery-region PSNR from 18.87 to 19.96\,dB and SSIM~\cite{wang2004image} from 0.5831 to 0.6029.
LPIPS~\cite{zhang2018unreasonable} decreases from 0.6069 to 0.5758, but remains above the Telea~\cite{telea2004image} baseline's 0.5466.

We assess the geometric effects separately on Bonn.
On the 20-video, ten-view evaluation, \armfilm{} has AbsRel 0.0257, compared with 0.0260 and 0.0258 for \armrgb{} at loss weights $\lambda=0.05$ and $\lambda=0.5$.
Under the same ten-view protocol, \armgeo{} with photometric loss weight $0.3$ has background AbsRel 0.0264, compared with 0.0257 for \armfilm{}.
These depth results use the same 20-video, fixed-ten-target protocol and depth-scale fitting as \cref{tab:bonn-exclusion}.
We additionally compare foreground leakage on the 22-video, ten-view protocol described in \cref{sec:supp-mechanism-analysis}, reporting median paired differences as \armrgb{} minus \armfilm{}.
The differences are $+0.000058$ (95\% CI: $[-0.00192,0.00182]$) at $\lambda=0.05$ and $-0.000870$ ($[-0.00602,0]$) at $\lambda=0.5$.
Both percentile intervals include zero and use 5,000 video-level bootstrap resamples.
Across the RGB and GeoPhoto variants, auxiliary appearance supervision yields no consistent additional geometric benefit over the corresponding conditioning-only models in the evaluated configurations.
We therefore retain the geometric training objective and post-hoc color transfer in the main configuration, and examine the RGB head's ability to fill the remaining appearance gaps below.

\paragraph{Qualitative appearance completion.}
\Cref{fig:supp-rgb-completion-iphone} shows \armrgb{} with $\lambda=1.5$ on the same eight iPhone scenes, input views, and exclusion masks as \cref{fig:supp-iphone-static-exclusion}.
For visualization, we retain geometrically transferred colors with local color correction wherever valid source colors are available and use the RGB head predictions in the remaining masked regions.
Although trained on cross-view-supported masked pixels, the head also supplies colors in regions that remain gray after transfer in \cref{fig:supp-iphone-static-exclusion}.
The examples show that the completion head can extend surrounding surface colors into these regions, while fine texture remains blurred and some color mismatches remain visible.
The head predicts these colors from learned reconstruction features without an additional generative model.

{\small
\makeatletter\@startbibunitorrelax\makeatother

}
\clearpage
\end{bibunit}

\end{document}